\documentclass{article}

\PassOptionsToPackage{numbers, compress}{natbib}
 \usepackage[preprint]{neurips_2026}

\usepackage[utf8]{inputenc} 
\usepackage[T1]{fontenc}    
\usepackage{hyperref}       
\usepackage{url}            
\usepackage{booktabs}       
\usepackage{amsfonts}       
\usepackage{nicefrac}       
\usepackage{microtype}      
\usepackage{xcolor}         
\usepackage{subcaption}
\usepackage{listings}
\usepackage{tabularx}

\usepackage{tcolorbox}
\tcbuselibrary{breakable}
\usepackage{pifont}
\usepackage{graphicx}
\usepackage{placeins}
\usepackage{amsmath}
\usepackage{wrapfig}
\usepackage{enumitem}

\newcommand{\cmark}{{\color{green!60!black}\ding{51}}} 

\title{\textsc{ShareChat}: A Dataset of Chatbot \\Conversations in the Wild}

\author{
Yueru Yan \quad Tuc Nguyen \quad Bo Su \quad Melissa Lieffers \quad Thai Le \\
Indiana University \\
Bloomington, USA\\
\texttt{\{yueryan,tucnguye,subo,mealieff,tle\}@iu.edu}
}

\begin{document}

\maketitle

\begin{abstract}
By evaluating Large Language Models (LLMs) through uniform, text-only 
interfaces, current academic benchmarks obscure how the unique designs and affordances of distinct commercial platforms shape real-world user behavior and system performance.
To bridge this gap, we present \textsc{ShareChat}, the first large-scale corpus of 142,808 conversations (660,293 turns) collected from publicly shared URLs on ChatGPT, Perplexity, Grok, Gemini, and Claude. \textsc{ShareChat} preserves native platform affordances, including citations, thinking traces, and code artifacts, across 95 languages and the period from April 2023 to October 2025, complementing existing corpora that homogenize these interactions. To demonstrate the dataset's evaluative utility, we present three case studies: a conversation completeness analysis assessing cross-platform differences in intent satisfaction, a source grounding analysis comparing citation strategies between search-augmented systems, and a temporal analysis revealing divergent response latency dynamics. Together, these analyses demonstrate research questions that are inaccessible to single-platform or stripped-affordance corpora. The dataset is publicly available.
\footnote{\url{https://huggingface.co/datasets/tucnguyen/ShareChat}}

\end{abstract}

\section{Introduction}
\label{sec:intro}
Conversational Large Language Model (LLM)-based chatbot services have evolved rapidly in the past three years. The first widely adopted general-purpose LLM chatbot, ChatGPT, was launched in November 2022 and has reached more than 700 million weekly active users by mid-2025 \cite{openai2025usagepatterns}. Shortly after, in late 2022, Perplexity emerged as an answer engine combining conversational interaction with web search \cite{perplexityblog2023}. Following this initial wave, 2023 saw a rapid industry expansion: Anthropic introduced the Claude family \cite{Anthropic_2023, anthropic2024claude3card, Anthropic_2025}, Google deployed the Gemini series \cite{gemini2023report}, and xAI launched Grok, integrating it directly into the social media platform X \cite{xai2023grok}.

\begin{table*}[tb!] 
\centering  
\footnotesize
\setlength{\tabcolsep}{3pt}
\caption{Comprehensive dataset comparison showing existing corpora, multi-platform aggregate, and per-platform breakdown. Our multi-platform dataset demonstrates superior linguistic diversity (95 languages) and conversation depth. Token statistics computed using Llama-2 tokenizer.}
\label{table:comprehensive_comparison}
\resizebox{\textwidth}{!}{%
\begin{tabular}{lccccccc} 
    \toprule 
    \textbf{Dataset} & \textbf{\#Convs} & \textbf{\#Users} & \textbf{\#Turns} & \textbf{\#Avg. Turns} & \textbf{\#Avg. User Tok} & \textbf{\#Avg. Chatbot Tok} & \textbf{\#Langs} \\ 
    \midrule 
    \multicolumn{8}{l}{\textit{Existing Public Datasets}} \\ 
    Alpaca & 52,002 & -- & 52,002 & 1.00 & $19.67_{\pm 15.19}$ & $64.51_{\pm 64.85}$ & 1 \\ 
    Open Assistant & 46,283 & 13,500 & 108,341 & 2.34 & $33.41_{\pm 69.89}$ & $211.76_{\pm 246.71}$ & 11 \\ 
    Dolly & 15,011 & -- & 15,011 & 1.00 & $110.25_{\pm 261.14}$ & $91.14_{\pm 149.15}$ & 1 \\ 
    ShareGPT & 94,145 & -- & 330,239 & 3.51 & $94.46_{\pm 626.39}$ & $348.45_{\pm 269.93}$ & 41 \\ 
    LMSYS-Chat-1M & 1,000,000 & 210,479 & 2,020,000 & 2.02 & $69.83_{\pm 143.49}$ & $215.71_{\pm 1858.09}$ & 65 \\ 
    WILDCHAT &3,199,860 &1,833,730 & 5,167,182 &1.61 & $1993.31_{\pm 8800.18}$ & $519.73_{\pm 830.63}$ & 76 \\
    \midrule 
    \multicolumn{8}{l}{\textit{Our Multi-Platform Dataset}} \\ 
    \textbf{Multi-Platform (Total)} & \textbf{142,808} & -- & \textbf{660,293} & \textbf{4.62} & $135.04_{\pm 1820.88}$ & $1,115.30_{\pm 1764.81}$ & \textbf{95} \\ 
    \midrule 
    \multicolumn{8}{l}{\textit{Per-Platform Breakdown}} \\   
    ChatGPT & 102,740 & -- & 542,148 & 5.28 & $142.35_{\pm 1191.57}$ & $1,230.25_{\pm 2448.38}$ & 84 \\ 
    Perplexity & 17,305 & 4,763 & 24,378 & 1.41 & $33.07_{\pm 261.74}$ & $573.33_{\pm 932.90}$ & 47 \\ 
    Grok & 14,415 & -- & 53,094 & 3.69 & $179.04_{\pm 6999.90}$ & $1,141.74_{\pm 1506.97}$ & 54 \\ 
    Gemini & 7,402 & -- & 36,422 & 4.92 & $184.62_{\pm 1571.62}$ & $803.23_{\pm 1609.27}$ & 41 \\
    Claude & 946 & -- & 4,251 & 4.49 & $138.67_{\pm 2213.46}$ & $576.16_{\pm 1649.61}$ & 18 \\
    \bottomrule
\end{tabular}}
\end{table*}

Although all of these services are built on text-based LLMs, they differ in interface design, supported features, and safety policies, which in turn shape how users interact with them. For example, Grok can surface live posts from X when providing answers, Claude models are optimized and evaluated for coding, math, and analysis tasks \cite{Anthropic_2023, anthropic2024claude3card}, and Perplexity consistently presents responses with explicit source citations \cite{perplexityhelp2024}. Commercial systems typically undergo continuous reinforcement learning and fine tuning on conversation logs and human feedback \cite{ouyang2022instructgpt,bai2022hh,chen2024chatgpt}, which tends to reinforce platform-specific strengths and norms over time. 


However, current research often fails to capture the complexity of real-world LLM deployment.  
While commercial platforms have evolved into complex ecosystems with unique features, the scientific community relies on datasets that homogenize these interactions, whether in synthetic~\cite{xu2024magpie, li2025infinity, ding2023enhancing} or real-world datasets including WildChat~\cite{zhao2024wildchat}, ShareLM~\cite{don2025sharelm}, LMSYS-Chat-1M~\cite{zheng2023lmsys}, OpenAssistant~\cite{kopf2023openassistant}, Alpaca~\cite{taori2023stanford}, ShareGPT\footnote{\url{https://sharegpt.com/}}, and Dolly~\cite{conover2023free}.
Specifically, WildChat \cite{zhao2024wildchat} compiles millions of conversations between users and ChatGPT collected via an access gateway that offered free usage in exchange for consent to share logs. LMSYS-Chat-1M \cite{zheng2023lmsys} contains one million conversations with twenty-five state of the art models recorded through a single web interface (the Vicuna demo and Chatbot Arena). OpenAssistant Conversations \cite{kopf2023openassistant} provides human generated assistant style dialogues and detailed preference annotations to support alignment research. 
These datasets have several limitations that restrict their utility for training and evaluating modern LLM systems:

\begin{enumerate}[leftmargin=*, noitemsep, topsep=0pt]
    \item \textbf{Artificial Uniformity:} Existing datasets typically collect data via a single, generic interface to standardize the process, which distorts reality. By forcing every model into the same neutral interface, these datasets strip away the unique tools that guide user behavior as they do not interact with a generic ``model'' in the real world, but a specific product. A corpus preserving these distinctions would enable stratified evaluation across product ecosystems.
    \item \textbf{Loss of Non-Textual Context:} In addition, while current datasets are largely limited to plain text, authentic LLM interactions rely on rich structural elements, such as visible thinking traces and embedded source citations, that are often stripped from standard logs.
    This prevents researchers from studying how critical product features actually influence user behavior and prompting strategies. Retaining such elements would provide supervision signals for training and evaluating structured generation capabilities.
    \item \textbf{Limited Interaction Depth:} Moreover, although some datasets explicitly include multi-turn conversations \cite{zhao2024wildchat}, the average depth of a conversation is still relatively short. In practice, the use of conversational LLMs is much more diverse and complex: users ask follow-up questions, change goals, or co-construct content over many turns. Recent work has shown that even state of the art models often get ``lost'' in multi-turn conversations, with reliability degrading as instructions are refined across turns \cite{laban2025llmslostmultiturnconversation}. Longer real user-LLM conversations, extended in both turn count and token length, are essential for benchmarking multi-turn reliability and for providing training signals that capture how user goals evolve across turns.
    \item \textbf{``Forced'' Sharing:} Finally, most datasets rely on participants who know they are being watched, introducing the \textbf{Observer Bias} (or Hawthorne Effect) \cite{mccambridge2014hawthorne}. Users who are aware they are being monitored for research purposes may consciously or unconsciously alter their behavior, which often exhibits higher social desirability compared to authentic, unobserved private usage. 
\end{enumerate}

In this study, we address these gaps by constructing a large scale, multi-platform corpus of authentic user-LLM conversations.
We compiled \textbf{142,808 conversations} with more than \textbf{660,293 interaction turns} from five widely used platforms: ChatGPT, Perplexity, Grok, Gemini, and Claude, as shown in Table~\ref{table:comprehensive_comparison}. All conversations were collected from URLs that users chose to share publicly on the respective platforms, spanning more than thirty months of chatbot usage (April 2023 to October 2025) and covering 95 distinct languages. As \textsc{ShareChat} relies on \textit{post-hoc} sharing, it therefore captures authentic usage patterns from a diverse, global user base, rather than simulated prompts or tasks designed by researchers. Each conversation record contains the complete sequence of user and assistant turns, along with platform-specific metadata, when available, timestamps, model thinking traces, or source links attached by the interface. Table~\ref{table:features} outlines the distinct data attributes captured across the five platforms, and the details are shown in Appendix~\ref{ssec:platform}. Data collection is done with IRB approval.
\begin{wraptable}{r}{0.5\textwidth} 
    \footnotesize
    \setlength{\tabcolsep}{0.75pt}
    \centering
    \caption{Feature overview and data extraction capabilities across platforms. Checkmarks indicate feature presence.}
    \label{table:features}
    \begin{tabular}{lccccc}
        \toprule
        \textbf{Feature/Field} & 
        \rotatebox{45}{\textbf{ChatGPT}} & 
        \rotatebox{45}{\textbf{Perplexity}} & 
        \rotatebox{45}{\textbf{Grok}} & 
        \rotatebox{45}{\textbf{Gemini}} & 
        \rotatebox{45}{\textbf{Claude}} \\
        \midrule
        Textual Content & \cmark & \cmark & \cmark & \cmark & \cmark \\
        \midrule
        Source Citations & - & \cmark & \cmark & - & - \\
        Thinking Blocks  & - & -      & \cmark & - & \cmark \\
        Code Artifacts   & - & -      & -      & - & \cmark \\
        Analysis Blocks  & - & -      & -      & - & \cmark \\
        \midrule
        Turn Timestamps  & \cmark & - & \cmark & - & - \\
        Model Version    & \cmark & - & \cmark & \cmark & - \\
        View/Share Counts& - & \cmark & - & - & - \\
        \bottomrule
    \end{tabular}
\end{wraptable}

To demonstrate the specific utility of \textsc{ShareChat}, besides describing the basic statistics of the corpus, we present three illustrative case studies. First, we perform conversation completeness analysis to assess if user intentions are fully addressed, leveraging \textsc{ShareChat}'s high interaction depth to capture complex goal resolutions. Second, we conduct source domain analysis to evaluate how models ground claims, utilizing preserved non-textual metadata to study citation behaviors. Third, we execute temporal analysis using turn-level timestamps, to reveal authentic interaction rhythms and engagement patterns.


Our contributions are threefold. First, we construct the first share-URL-based corpus that preserves native interaction affordances across five platforms simultaneously. The post-hoc public-sharing access model is complementary to consent-based gateways such as WildChat~\cite{zhao2024wildchat} and opt-in plugins such as ShareLM~\cite{don2025sharelm}. Second, the corpus carries substantially longer conversations on average, lower observed levels of toxic content, and broader linguistic coverage than prior corpora. Third, we demonstrate the dataset's utility through three case studies that reveal platform-dependent patterns inaccessible to existing corpora.

\section{Data Collection}
\label{sec:data_collection}
For the five different platforms, we develop systematic, platform-specific data extraction pipelines to collect conversation content, metadata, and platform-specific features from publicly accessible shared conversation URLs found on the Internet. Although we find that such URLs are often shared on social media platforms such as X (formerly Twitter), Reddit, and Discord, exhaustively monitoring these individual ecosystems presents significant scalability challenges. To tackle this, we utilize Internet archival services, specifically the Internet Archive (Wayback Machine), to search for and extract URLs matching specific patterns (e.g., \textit{``chatgpt.com/share/*''} for ChatGPT). This archival-based discovery method is a well-established strategy for aggregating public digital resources, previously employed to compile datasets for web-scale text \cite{raffel2020t5}, historical news archives \cite{hamborg2019automated}, and corporate disclosure analysis \cite{boulland2026}. For each platform, data collection covers a distinct time window, as summarized in Appendix~\ref{ssec:timeframe_and_link}. Beyond the snapshot reported in this paper, we are actively maintaining the collection pipeline and continuously ingesting newly shared conversations as they appear in the wild, supporting longitudinal study of platform evolution and progressive coverage of under-represented systems and new platforms as they emerge.

For every shared conversation page, we use an automated browser control with Selenium to interact with the page, retrieve the rendered HTML, and parse it into a structured JSON representation. The resulting records include the ordered sequence of conversation turns, user prompts, model responses, and platform-specific metadata. When certain elements are only visible after user interaction, we script the corresponding actions. For example, to collect the thinking content of Claude, we first locate the thinking bar element and then trigger a click on it when the content is only accessible through interaction. 


\paragraph{Filtering of Identifiable Information.} We prioritize user privacy by anonymizing personally identifiable information (PII). We use Microsoft's Presidio\footnote{\url{https://microsoft.github.io/presidio/}} as the framework to remove PII across various data types, such as names, phone numbers, emails, credit cards, driver licenses and URLs, in multiple languages. To ensure rigorous privacy preservation, we employed an automated LLM-based evaluator to audit the PII removal results, with details provided in Appendix~\ref{sssec:pii}.

\section{Preliminary Analysis}
The final dataset contains conversations from five platforms, featuring a wide range of languages, diverse user prompts and longer conversations compared to other datasets. In this section, we describe the basic statistics of the dataset, and the results of our toxicity analysis and topic distribution.

\subsection{Basic Statistics}

In total, \textsc{ShareChat} has \textbf{142,808 conversations} with more than \textbf{660,293 interaction turns}. Table~\ref{table:comprehensive_comparison} compares \textsc{ShareChat} with existing corpora across key metrics: turn count, token length, and language diversity. The language detection was performed at message-level using lingua-py and then combined into conversation-level statistics. Token counts were computed using the Llama-2 tokenizer for consistent cross-platform comparison \cite{zhao2024wildchat}.

\begin{figure}[htb!]
    \centering
    \begin{minipage}{0.48\textwidth}
        \centering
        \includegraphics[width=\linewidth]{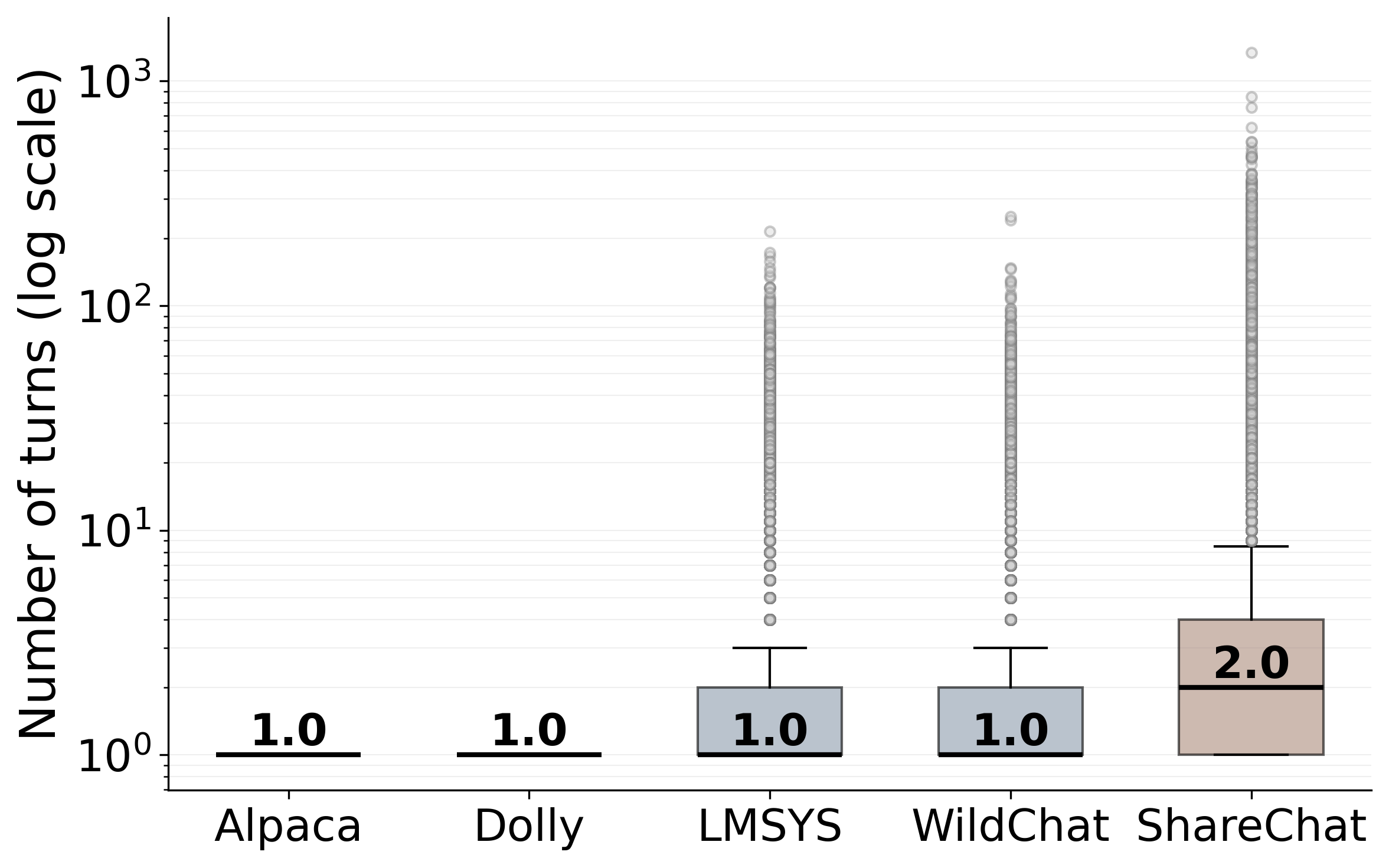}
        \caption{Turn length distribution across datasets (log scale). \textsc{ShareChat} exhibits a higher median turn count (2.0) compared to Alpaca, Dolly, LMSYS, and WildChat (all 1.0).}
        \label{fig:turns_public_comparison}
    \end{minipage}
    \hfill
    \begin{minipage}{0.48\textwidth}
        \centering
        \includegraphics[width=\linewidth]{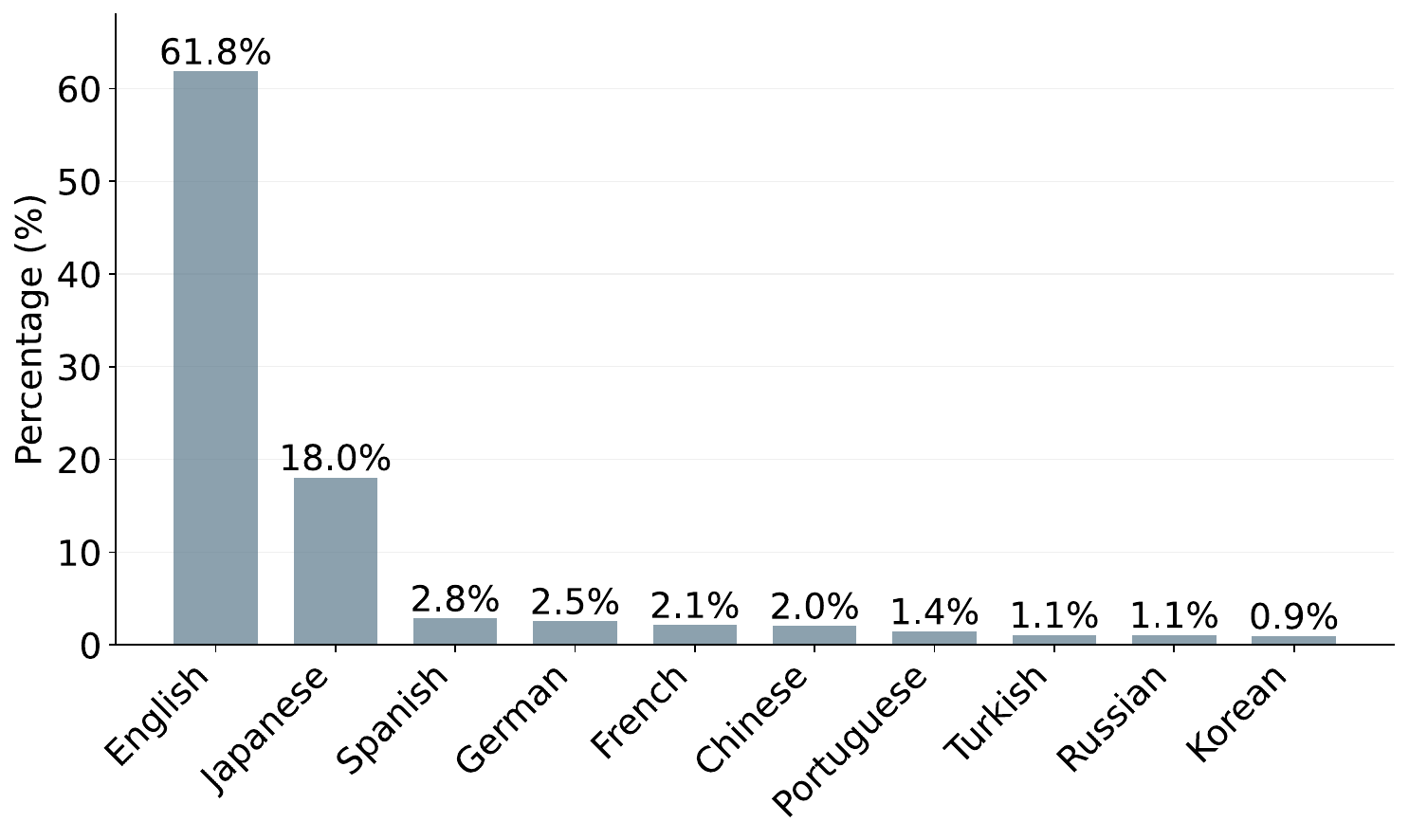}
        \caption{Distribution of the top 10 languages in \textsc{ShareChat}. English is the dominant language (61.8\%), followed by Japanese (18.0\%), with the remaining languages each accounting for less than 3\%.}
        \label{fig:lang_agg}
    \end{minipage}
\end{figure}

\textbf{\textsc{ShareChat} distinguishes itself by extending the scope of analysis along two critical dimensions: interaction depth and content volume.} First, conversations are substantially longer, averaging 4.62 turns with a heterogeneous distribution that consistently exceeds existing public corpora (Figure~\ref{fig:turns_public_comparison} and Appendix~\ref{ssec: turn_distri}). Second, model responses are significantly denser, with a mean of 1,115.30 tokens compared to just 519.73 for WildChat, suggesting the presence of richer and more substantive problem-solving content from the LLM side.

\textsc{ShareChat} also demonstrates greater linguistic diversity by spanning 95 distinct languages, whereas the most diverse competitor covers only 76. For languages covered, as shown in Figure~\ref{fig:lang_agg}, English accounts for the clear majority of conversations across all platforms, with Japanese as the second most frequent language and all other languages each contributing a much smaller share. Compared to previous datasets, \textsc{ShareChat} has a more balanced language distribution.




\FloatBarrier
\subsection{Toxicity Analysis}
\label{sec:toxicity_analysis}
To assess toxicity in \textsc{ShareChat}, we employ both Detoxify \cite{hanu2020detoxify} and OpenAI Moderation API\footnote{\url{https://platform.openai.com/docs/guides/moderation}} to detect toxic content in the conversations, following the approach used in WildChat~\cite{zhao2024wildchat}. Detoxify is a pre-trained multilingual toxicity classification model that computes toxicity scores across seven dimensions, while OpenAI Moderation provides commercial-grade content filtering. Given the language coverage limitations of Detoxify, we retain only conversational turns in languages explicitly supported by the model, with more details shown in Appendix~\ref{ssec: toxicity_results}. For Detoxify, we apply the same classification threshold as WildChat, marking messages as toxic when the toxicity score exceeds 0.1, while OpenAI Moderation uses its internal flagging mechanism to identify toxic content.

\begin{wraptable}{r}{0.55\textwidth}
\centering
    \caption{Turn-level toxicity rates (\%) by platform, measured by Detoxify and OpenAI Moderation API. Toxicity rates for user messages are generally higher than for LLM responses.}
    \label{tab:turn_level_toxicity}
    \small
    \begin{tabular}{lcccc}
    \toprule
    & \multicolumn{2}{c}{\textbf{LLM Turns}} & \multicolumn{2}{c}{\textbf{User Turns}} \\
    \cmidrule(lr){2-3} \cmidrule(lr){4-5}
    \textbf{Platform} & Detoxify & OpenAI & Detoxify & OpenAI \\
    \midrule
    ChatGPT    & 1.4 & 2.9 & 4.1 & 2.8 \\
    Perplexity & 0.5 & 1.8 & 2.8 & 1.3 \\
    Grok       & 3.7 & 6.3 & 5.2 & 4.0 \\
    Gemini     & 1.3 & 3.7 & 4.3 & 3.8 \\
    Claude     & 6.4 & 3.1 & 5.6 & 2.7 \\
    \midrule
    All & 1.6 & 3.2 & 4.1 & 2.9\\
    \bottomrule
\end{tabular}
\end{wraptable}

The results for turn-level toxicity are shown in Table~\ref{tab:turn_level_toxicity}, which reveal substantial differences between the two detection methods, consistent with limited agreement observed in prior work \cite{zhao2024wildchat}.

To examine whether platforms ``mirror'' user toxicity, we test within each platform at the conversation level: for each conversation we compute the mean toxicity score over user turns and over LLM turns, then correlate these two per-conversation means across all conversations in the platform. All five platforms show a moderate-to-strong positive Spearman correlation under both detectors (Detoxify: $\rho = 0.45$–$0.66$; OpenAI: $\rho = 0.45$–$0.63$; all Bonferroni-corrected $p < 10^{-65}$, full per-platform numbers in Appendix~\ref{ssec: toxicity_results}. Conversations with more toxic users thus elicit more toxic LLM responses within every platform we measured. It is also worth noting that Claude's per-platform rates rest on a smaller subsample and carry wider uncertainty than the larger platforms.

Most importantly, \textbf{\textsc{ShareChat} exhibits lower toxicity rates compared to existing benchmarks}. Our overall user toxicity rates of 2.9\% are considerably lower than the comparable figures reported in WildChat's Table 7~\cite{zhao2024wildchat} which records user toxicity of 6.05\% on its own corpus and re-evaluates LMSYS-Chat-1M at 3.08\% using the same OpenAI Moderation pipeline. Similarly, our LLM toxicity rates of 3.2\% via OpenAI are lower than the 5.18\% reported by WildChat and comparable to the 4.12\% in LMSYS-Chat-1M.

\FloatBarrier


\subsection{Topic Analysis}
\label{sec:topic_analysis}

\begin{wrapfigure}{r}{0.45\textwidth}
    \centering
    \includegraphics[width=\linewidth]{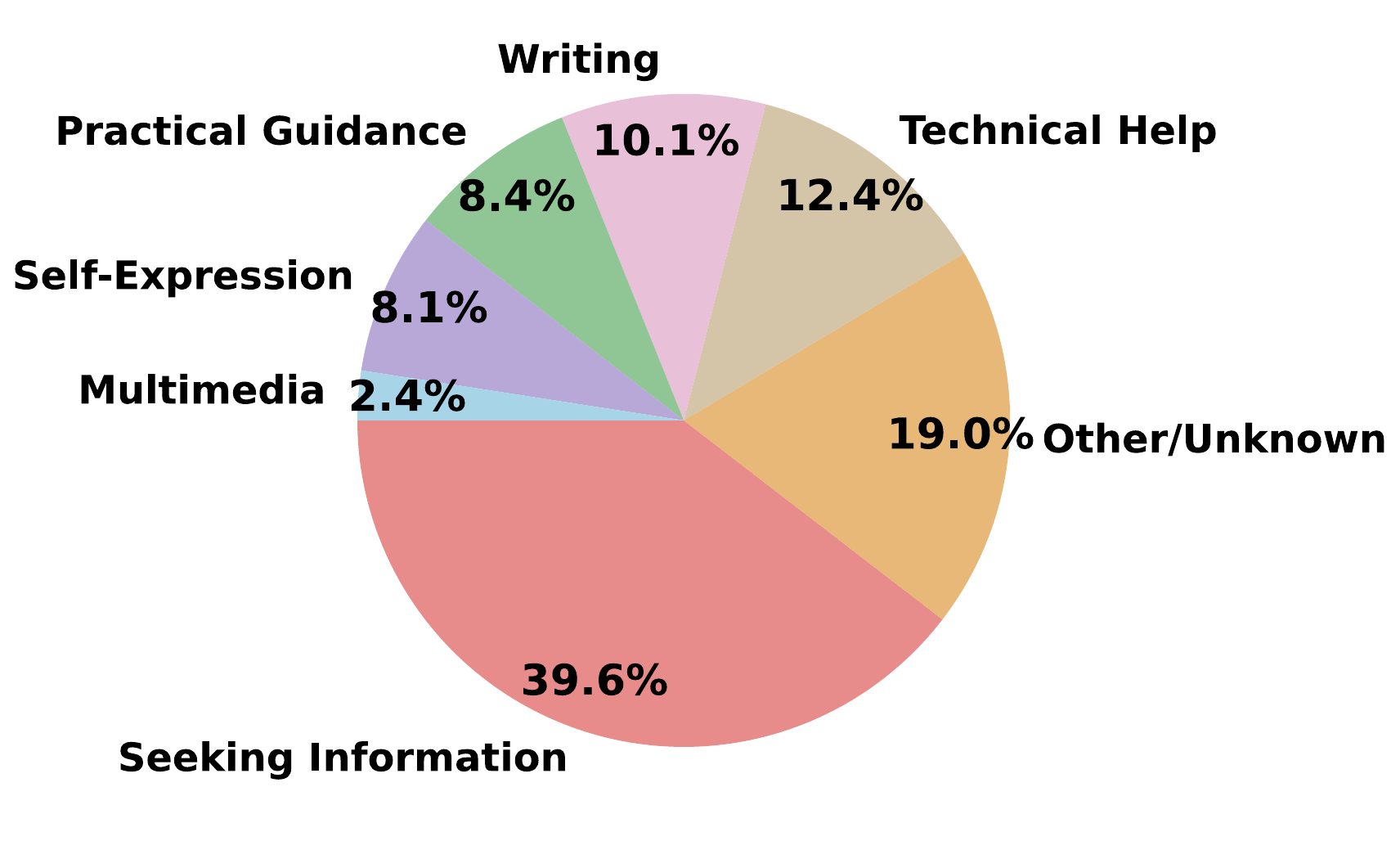}
    \caption{Average topic distribution of user requests across five platforms. Seeking Information is the most prevalent intent (39.6\%), followed by Other/Unknown (19.0\%) and Technical Help (12.4\%).}
    \label{fig:avg_topic_discovery}
\end{wrapfigure}

We classify each user message into one of 24 predefined topic categories using multilingual LLMs Llama-3.1-8B-Instruct \footnote{https://huggingface.co/meta-llama/Llama-3.1-8B-Instruct} under a few-shot prompting setup. To enable large-scale processing, we run
inference with 4-bit quantization, batch size of 32. Model outputs are normalized through a post-processing pipeline to handle formatting variants and stray meta-responses, ensuring reliable mapping to our fixed taxonomy. To validate the classifier, two authors independently labeled 100 English user messages sampled from the five platforms. Inter-annotator agreement was substantial (Cohen's $\kappa = 0.701$); after adjudicating disagreements into a gold-standard label set, the LLM classifier achieved $82.0\%$ accuracy and a macro-F1 of $76.9\%$ against the gold-standard labels. For downstream analyses, the 24 fine-grained categories are further consolidated into seven higher-level groups to support both granular and coarse-grained comparisons across platforms. The seven high-level categories of user requests are plotted in Figure~\ref{fig:avg_topic_discovery}. Overall, the averaged distribution reveals that information seeking remains the core driver of LLM interactions, with creative, technical, and advisory uses forming important but secondary dimensions of user behavior. 

Beyond the aggregate distribution shown in Figure~\ref{fig:avg_topic_discovery}, the per-platform breakdown reveals that \textbf{users
align their requests with each platform's perceived strengths rather
than treating chatbots as interchangeable assistants}. Perplexity is
overwhelmingly concentrated on Seeking Information (63.3\%), consistent
with its search-oriented design, with minimal Self-Expression (1.2\%)
and Multimedia (0.7\%) usage. Claude stands apart with the highest
share of Technical Help (17.0\%). ChatGPT
(34.2\%), Gemini (38.3\%), and Grok (42.8\%) show progressively higher
Seeking Information shares, while Writing stays stable at
$\approx 10$--$11\%$ across the three. Grok exhibits the highest
Multimedia usage (4.4\%), potentially reflecting its integration with
media-rich X content. These platform-level differences indicate that
users treat each service as a specialized tool, supporting the
platform-aware evaluation framing of this paper. Per-platform pie
charts, prompts, category definitions, and full
classification statistics are provided in Appendix \ref{detail_user_request_topic_distributions}.

\section{Representative Analyses}

\subsection{Conversation Completeness Analysis}
\label{sec:conversation_completeness_analysis}
To assess how comprehensively LLMs fulfill user needs in multi-turn dialogues, we develop an automated completeness analysis pipeline that evaluates whether each user intention across a conversation is adequately addressed by the assistant. Our analysis reveals distinct performance across platforms, with ChatGPT and Claude demonstrating the highest consistency in fully satisfying user intentions, while Gemini, Grok and Perplexity exhibit significantly wider variance in conversational quality across multi-turn interactions.

\begin{wrapfigure}{r}{0.5\textwidth}
    \centering
    \includegraphics[width=\linewidth]{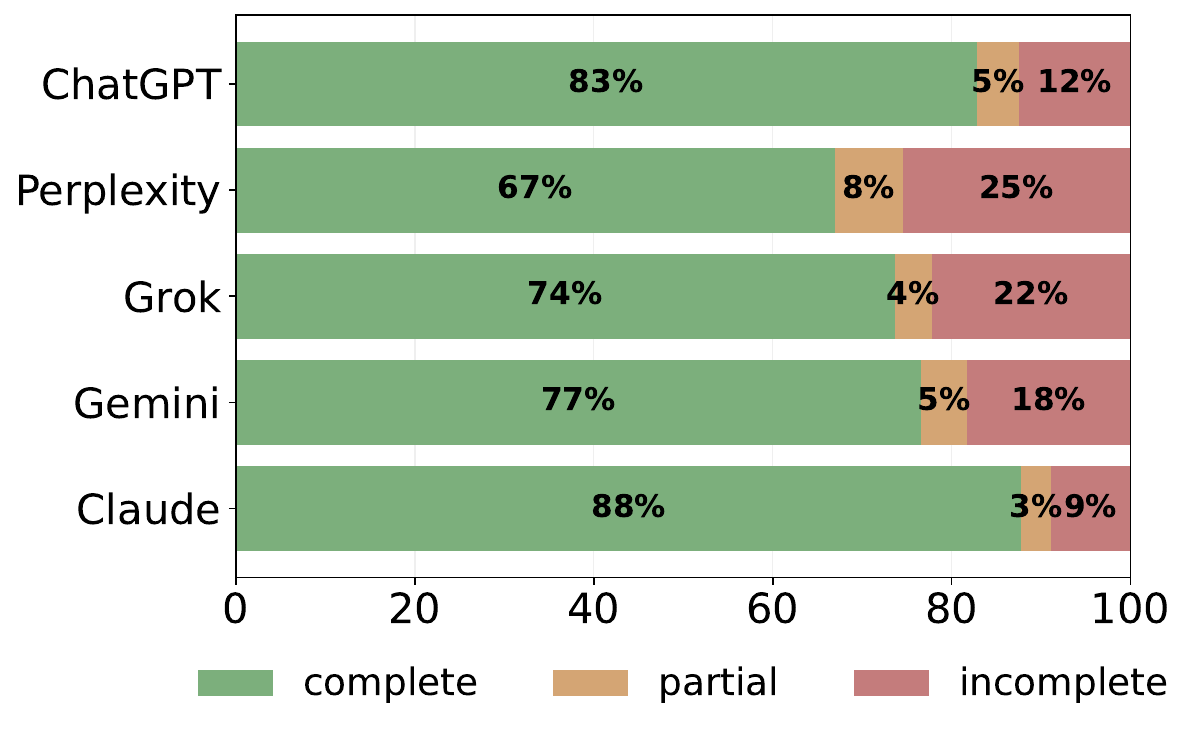}
    \caption{Conversation completeness by platform, measured as the proportion of user intentions receiving complete, partial, or incomplete verdicts. Claude achieves the highest full-completion rate (88\%), followed by ChatGPT (83\%) and Gemini (77\%), while Perplexity shows the lowest (67\%).}
    \label{fig:completeness_verdicts}
\end{wrapfigure}

To assess the conversation completeness, we implement a three-stage pipeline using Qwen3-8B~\cite{yang2025qwen3}. First, we extract distinct user intentions from the conversation history. Second, we classify the resolution of each intention as \textit{complete}, \textit{partial}, or \textit{incomplete}. Finally, we compute a conversation-level completeness score by weighting these verdicts. We sample 100 conversations (20 per platform) for human evaluation. 
Three authors of this paper independently evaluated 133 intentions derived from 64 English conversations out of the 100 sampled conversations. For intention extraction, annotators agreed that among the intentions extracted by the LLM, 96.24\% were correct (raw three-way agreement 85.83\%; pairwise Cohen's $\kappa = 0.21$–$0.53$, attenuated by skewed marginals since most intentions were judged correct). For intention fulfillment, using the post-discussion consensus labels as our ground truth, the LLM achieved an intention fulfillment accuracy of 87.40\% and an F1-score of 89.90\%.

Through this annotation scheme, the number of extracted intentions per conversation varies systematically across platforms, with ChatGPT and Claude conversations containing a median of 2 intentions, while Gemini, Grok, and Perplexity show a median of 1 intention per conversation. Full prompts and implementation details are provided in Appendix~\ref{conversation_completeness_prompt}.

Different platforms reveal substantial variation in how effectively they address user intentions in multi-turn conversations. ChatGPT and Claude exhibit the highest rates of complete verdicts in Figure~\ref{fig:completeness_verdicts}, with both platforms fully resolving the majority of user intentions. Gemini and Grok show more balanced distributions between complete and partial verdicts, with a substantial proportion of intentions receiving incomplete verdicts. Perplexity demonstrates an intermediate pattern with the highest rate of partial verdicts across all platforms. 

These findings highlight the \textbf{dataset's structural diversity, spanning single-turn queries to complex, multi-intention sessions.} The prevalence of partial verdicts reveals that real-world goals often remain incompletely satisfied, which is a dynamic obscured by benchmarks prioritizing single-turn correctness. By providing these completeness labels, \textsc{ShareChat} enables the development of metrics aligned with user satisfaction and offers rich signals for reinforcement learning.



\subsection{Source Grounding Analysis}

In this analysis, we observe distinct information retrieval paradigms between the two search-enabled platforms. As detailed in 
Table~\ref{table:features}, external grounding is a core feature for both systems: the Grok dataset contains 14,415 conversations with 8,242 (57.18\%) including sources, and the Perplexity dataset includes 17,305 conversations with 8,545 (49.38\%) containing sources. To characterize retrieval behavior, we extract the top-level domain from each cited URL and aggregate citation counts by domain across all conversations per platform.

Grok exhibits a concentrated reliance on social data. Its source distribution is heavily skewed toward the X platform, which accounts for 40,624 references, nearly triple the citations for Wikipedia (12,507). The remaining top sources are predominantly legacy news outlets such as the New York Times, Forbes, and The Guardian. This pattern suggests a design optimized for real-time, social-media-driven context, potentially offering higher timeliness but introducing risks related to information stability and bias.

In contrast, Perplexity demonstrates a diversified, encyclopedic 
retrieval strategy. Its distribution is significantly less 
concentrated, with English Wikipedia as the leading source (2,919 
citations), followed by community forums like Reddit (1,642), 
authoritative scientific repositories like NIH (1,339), and 
professional platforms such as GitHub and LinkedIn. This breadth 
reflects a retrieval paradigm that prioritizes authority and topical coverage over recency. Furthermore, Perplexity routinely supports deeper research sessions with a broad distribution of citation counts, whereas Grok conversations typically cite fewer than 5 sources per interaction (see Appendix~\ref{appendix:response_source_analysis} for full source distributions).

Taken together, these findings demonstrate that the \textbf{two 
platforms employ fundamentally distinct information retrieval 
strategies, prioritizing different types of external evidence to 
ground their generated responses}.


\subsection{Timestamp Analysis}

Timestamp metadata is available for ChatGPT and Grok, covering nearly all turns (99.97\% for ChatGPT and 100\% for Grok). For the small fraction of missing timestamps in ChatGPT data, we employ linear interpolation between known points where feasible, and drop affected turns when interpolation was not possible. We verify that results remained stable across different handling strategies, confirming that the missing data does not bias our findings. After removing extreme outliers above the 99th percentile, we observe systematic differences between platforms. ChatGPT exhibits longer mean user response times than Grok (1,580 vs.\ 931 seconds), while producing shorter mean LLM response times (18.4 vs.\ 24.6 seconds). We caution that ChatGPT latency trends (May 2023–Aug 2025) conflate platform behavior with four model upgrades, whereas the Grok window (Dec 2024–Oct 2025) is more stable.

Then, we examine how response times evolve over the course of a conversation by correlating turn position with response latency. ChatGPT exhibits a significant negative correlation between turn index and LLM response time (Pearson $r = -0.238$; Spearman $\rho = -0.427$), indicating progressively faster responses as conversations lengthen, consistent with caching, adaptation, or context optimization effects. In contrast, Grok shows a significant positive correlation (Pearson $r = 0.315$; Spearman $\rho = 0.254$), suggesting increasing computational overhead as context accumulates. User response times show minimal linear association with turn position on both platforms.
These divergent temporal dynamics highlight architectural differences in how the two systems handle extended interactions.

\begin{figure}[tb!]
    \centering
    \begin{subfigure}[c]{0.4\textwidth}
        \centering
        \includegraphics[width=\columnwidth]{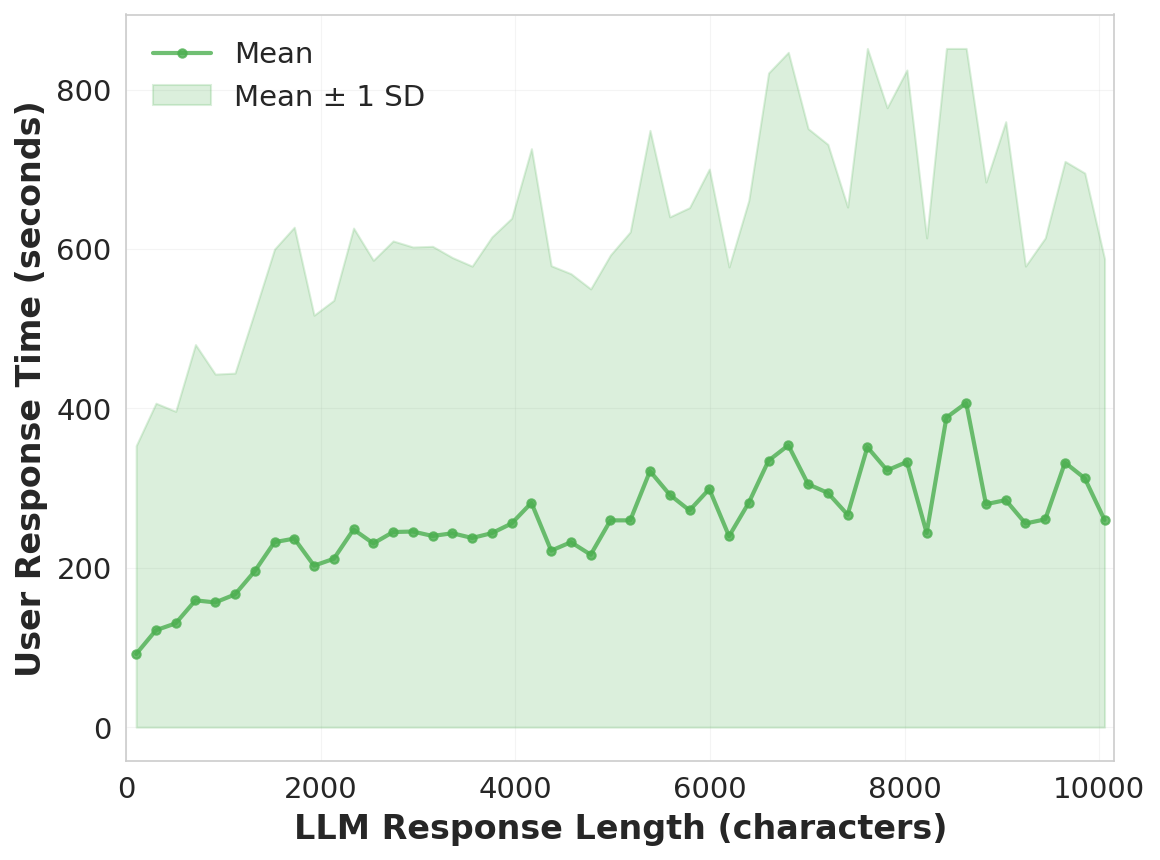}
        \caption{Grok}
        \label{fig:grok_source_dist}
    \end{subfigure}
    \hspace{0.1\textwidth}
    \begin{subfigure}[c]{0.4\textwidth}
        \centering
        \includegraphics[width=\columnwidth]{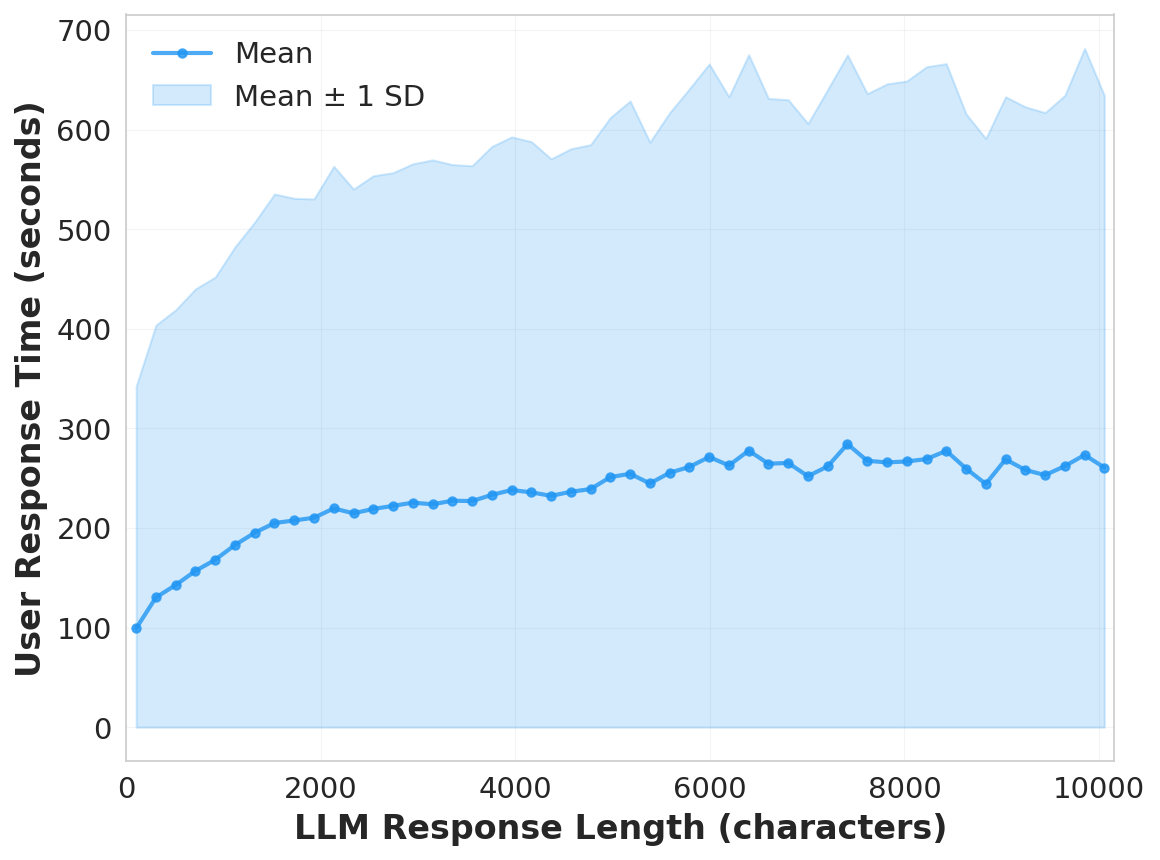}
        \caption{ChatGPT}
        \label{fig:perplexity_source_dist}
    \end{subfigure}
    \caption{Binned mean user response time as a function of LLM response length. Solid lines denote mean response time across users, while shaded regions indicate ±1 standard deviation.}
    \label{fig:grok_chatgpt_userbinned}
    \vspace{-5pt}
\end{figure}

To probe whether response length alone explains user processing time, we examine the relationship between LLM message length and subsequent user-response interval. Figure \ref{fig:grok_chatgpt_userbinned} shows that binned mean user-interval rises monotonically with LLM length, with the steepest growth at shorter outputs and a plateau beyond roughly 4–6k characters. However, the raw Pearson correlations are near zero on both platforms (ChatGPT r = 0.034; Grok r = 0.021; see Appendix~\ref{ssec: temporal}), reflecting the heavy-tailed dispersion of user-side intervals at every length bucket. Together, these results indicate that LLM response length predicts the conditional mean of user-side latency but explains very little of its individual-level variance, which is consistent with length being one factor among many, including 
task type, time of day, and user availability, rather than a 
dominant driver.








\section{Discussion}


The analyses above demonstrate several ways in which \textsc{ShareChat}
enables new research directions for the ML community.

\textbf{Within-platform extended-context dynamics.} 
\textsc{ShareChat}'s longer conversation contexts, averaging 4.62 turns 
compared to 2.02 in LMSYS-Chat-1M~\cite{zheng2023lmsys}, provide a critical 
resource for studying phenomena that emerge only over extended 
interactions. Prior work has shown that model reliability can degrade 
as instructions are refined across turns~\cite{laban2025llmslostmultiturnconversation, huang2026vulnerability}, yet most existing corpora offer limited opportunities to 
investigate such dynamics due to their predominantly single-turn 
structure. The presence of conversations extending into dozens of 
turns, combined with the diversity of intention counts per 
conversation, enables researchers to examine how user needs evolve, 
how models maintain coherence over long contexts, and when breakdowns 
occur. The prevalence of partial verdicts in our completeness 
analysis offers direct supervision for reinforcement learning from 
conversation-level feedback, where models must learn when to revisit 
unresolved intentions rather than simply optimizing single-turn 
accuracy. This signal is largely absent from existing benchmarks 
that prioritize binary correctness.

\textbf{Platform-Dependent Usage and Performance Variability.} 
Unlike the ``universal assistant'' assumption often found in general 
benchmarks, our cross-platform analysis reveals that user behavior 
is highly sensitive to the specific design and functional positioning 
of each platform. Our topic distribution analysis demonstrates that 
users align their requests with perceived platform strengths, such 
as using Perplexity for information retrieval versus Claude for 
technical help. This functional specialization is mirrored in 
conversation completeness: while ChatGPT and Claude exhibit narrow, 
concentrated completeness scores, Gemini and Grok show substantial 
variability. Notably, the high frequency of ``partial'' verdicts in 
Perplexity interactions reflects its distinct role as a search 
engine, where outputs often serve as intermediate steps rather than 
final answers. These platform-specific distributions provide natural 
experimental conditions for controlled comparison: researchers can 
use \textsc{ShareChat} to benchmark models against the actual task 
distributions each product faces, moving beyond uniform synthetic 
prompts to determine which features drive reliability for specific 
user intents.

\textbf{Rich Metadata for Downstream Tasks.} The preservation of 
rich metadata enables future research on retrieval-augmented 
generation and interaction pacing in naturalistic settings. Our 
source content analysis reveals pronounced differences in how 
platforms integrate external evidence: Grok relies overwhelmingly on 
concentrated source types, while Perplexity draws from a diversified 
set of domains. The preserved citation metadata provides a natural 
benchmark for evaluating RAG systems where researchers can assess 
citation accuracy, source diversity, and grounding quality directly 
from authentic system outputs rather than synthetic setups. 
Additionally, the turn-level timestamps enable latency-aware 
evaluation, a dimension absent from existing benchmarks. The 
contrasting dynamics between ChatGPT (decreasing latency over turns) 
and Grok (increasing latency) offer concrete test cases for studying 
how architectural choices affect user-facing performance in extended 
interactions.

We emphasize that \textsc{ShareChat} is complementary to, not a replacement for, existing corpora. The two access models embody different biases: WildChat's monitored gateway introduces observer bias~\cite{zhao2024wildchat}, whereas \textsc{ShareChat}'s post-hoc public sharing avoids observer effects but introduces self-selection bias, which is visible in our substantially lower toxicity rates and denser per-turn outputs. Neither distribution constitutes a ground truth, and treating the two as joint observations triangulates user-LLM interaction more completely than either alone. Beyond the three analyses demonstrated here, the dataset supports research directions including conversational breakdown detection, cross-platform transfer learning, and platform-aware evaluation frameworks. We encourage the community to explore these and other applications enabled by the dataset's unique combination of multi-platform coverage, rich metadata, and naturalistic multi-turn structure.

\section*{Limitations}

This work has four primary limitations. First, reliance on 
publicly shared URLs introduces self-selection bias: users who choose to share conversations likely filter for interactions they consider appropriate or valuable. While this mitigates the observer bias present in monitored collection datasets, it skews the corpus toward certain interaction types. Second, the dataset exhibits significant platform imbalance, with ChatGPT accounting for over 70\% of conversations and Claude less than 1\%. The skew partly reflects real-world share behavior and partly a sampling artifact we cannot disentangle. Per-platform collection windows also differ, so cross-platform comparisons partly conflate platform with collection era. Researchers should exercise caution when generalizing from minority platforms. Data collection is ongoing, and future releases will expand under-represented platforms. Third, our conversation completeness and topic analyses rely on LLM-based evaluation, which provides a scalable proxy for but not a perfect measure of user intent. Both human validations were conducted on English conversations, leaving performance on lower-resource languages unquantified. Finally, the specific analyses presented are illustrative rather than exhaustive; future work is needed to fully explore longitudinal dynamics as platform features evolve.

\section*{Ethical Considerations}

The \textsc{ShareChat} corpus consists exclusively of conversations from URLs that users voluntarily generated and shared via each platform's native sharing feature. We did not bypass authentication, scrape private accounts, or access non-public data. Data collection was conducted under IRB approval. To protect user privacy, we apply a rigorous de-identification pipeline: automated NER and regex matching redact PII across multiple languages, all user identifiers are cryptographically hashed, and the pipeline is validated through both automated LLM auditing and manual review (Appendix~\ref{sssec:pii}). All analyses reported in this paper rely on aggregated statistics, and no individual conversation is singled out in a way that could enable re-identification. We release \textsc{ShareChat} under a Creative Commons Attribution-NonCommercial 4.0 International (CC BY-NC 4.0) license for research use only, with the additional restriction that any de-anonymization or malicious surveillance is prohibited. Downstream users must also comply with each platform's original terms of service. \textsc{ShareChat} contains content that may be offensive or toxic, as quantified in Section~\ref{sec:toxicity_analysis}.

\section*{Acknowledgments}

\textbf{Methodological use of LLMs.} LLMs are used as analytical components of the pipeline reported in this paper: Llama-3.1-8B-Instruct for topic classification (Section~\ref{sec:topic_analysis}), Qwen3-8B for intention extraction and conversation-completeness evaluation (Section~\ref{sec:conversation_completeness_analysis}, Appendix~\ref{conversation_completeness_prompt}), and GPT-OSS-120B for PII-removal validation (Appendix~\ref{sssec:pii}). All LLM-generated outputs were verified through independent human annotation, cross-validation with multiple detection methods, or manual review of flagged cases, as detailed in the respective sections.

\textbf{Editorial use of LLMs.} LLMs were additionally used to assist with manuscript editing and proofreading. No LLM was used to generate substantive scientific claims, perform analyses, or synthesize results; all content was reviewed and verified by the authors for accuracy.

\bibliographystyle{unsrtnat} 
\bibliography{custom}

\clearpage

\appendix
\section{Appendix}

\subsection{Data collection}
\label{ssec:timeframe_and_link}
Table~\ref{table:platforms} details the target URLs used for extraction and the data collection timeframe for each platform. Creation timestamps were not captured for the Claude platform in this iteration; however, we plan to incorporate this metadata in future versions of the dataset.


\begin{table}[htb]
\centering
\caption{Supported AI chat platforms with share URL formats, primary design focus, and collection timeframes.}
\label{table:platforms}
\begin{tabular}{lll}
    \toprule
    \textbf{Platform} & \textbf{Share URL Format} & \textbf{Collection Period} \\
    \midrule
    ChatGPT & \texttt{chatgpt.com/share/*}  & May 2023 to Aug 2025 \\
    Perplexity & \texttt{perplexity.ai/search/*}  & Apr 2023 to Oct 2025 \\
    Grok & \texttt{grok.com/share/*}  & Dec 2024 to Oct 2025 \\
    Gemini & \texttt{gemini.google.com/share/*}  & Apr 2024 to Sep 2025 \\
    Claude & \texttt{claude.ai/share/*} & Not available \\
    \bottomrule
\end{tabular}
\end{table}

\subsection{Platform-Specific Interaction and Metadata Attributes}
\label{ssec:platform}
Different platforms also reflect distinct design philosophies, ChatGPT and Gemini offer general-purpose conversational assistance. Perplexity focuses on search-oriented information retrieval. Grok targets social media-integrated question answering. Claude specializes in technical and reasoning support. These structural differences align with intended representative analyses and shape the types of user interactions preserved in \textsc{ShareChat}. 

As shown in Table~\ref{table:features}, while ChatGPT and Gemini follow standard conversational formats, other platforms expose specialized content structures. Claude and Grok both expose 'thinking blocks' that surface intermediate reasoning. Claude additionally preserves versioned code artifacts and analysis blocks, enabling the study of iterative coding workflows. In contrast, Perplexity and Grok structure responses around retrieval. Perplexity organizes content into answers, sources, and images with inline citations, while Grok features a dedicated pane for web and social media sources (X posts). These structural differences allow for distinct lines of inquiry: technical workflow analysis for Claude versus citation accuracy and source integration studies for Perplexity and Grok.

The granularity of metadata also dictates the scope of possible temporal analysis. ChatGPT and Grok are the only platforms providing turn-level timestamps, which are essential for modeling conversation rhythm and response latency. ChatGPT further enriches this with model version identifiers (e.g., \texttt{gpt-4}). Conversely, Perplexity lacks turn-level timing but offers unique engagement metrics, including view counts, share counts, and last-updated timestamps, effectively treating conversations as evolving public artifacts rather than static logs. 

\subsection{Turn length Distribution}
\label{ssec: turn_distri}

\begin{figure}[tb!]
\centering
    \includegraphics[width=0.6\textwidth]{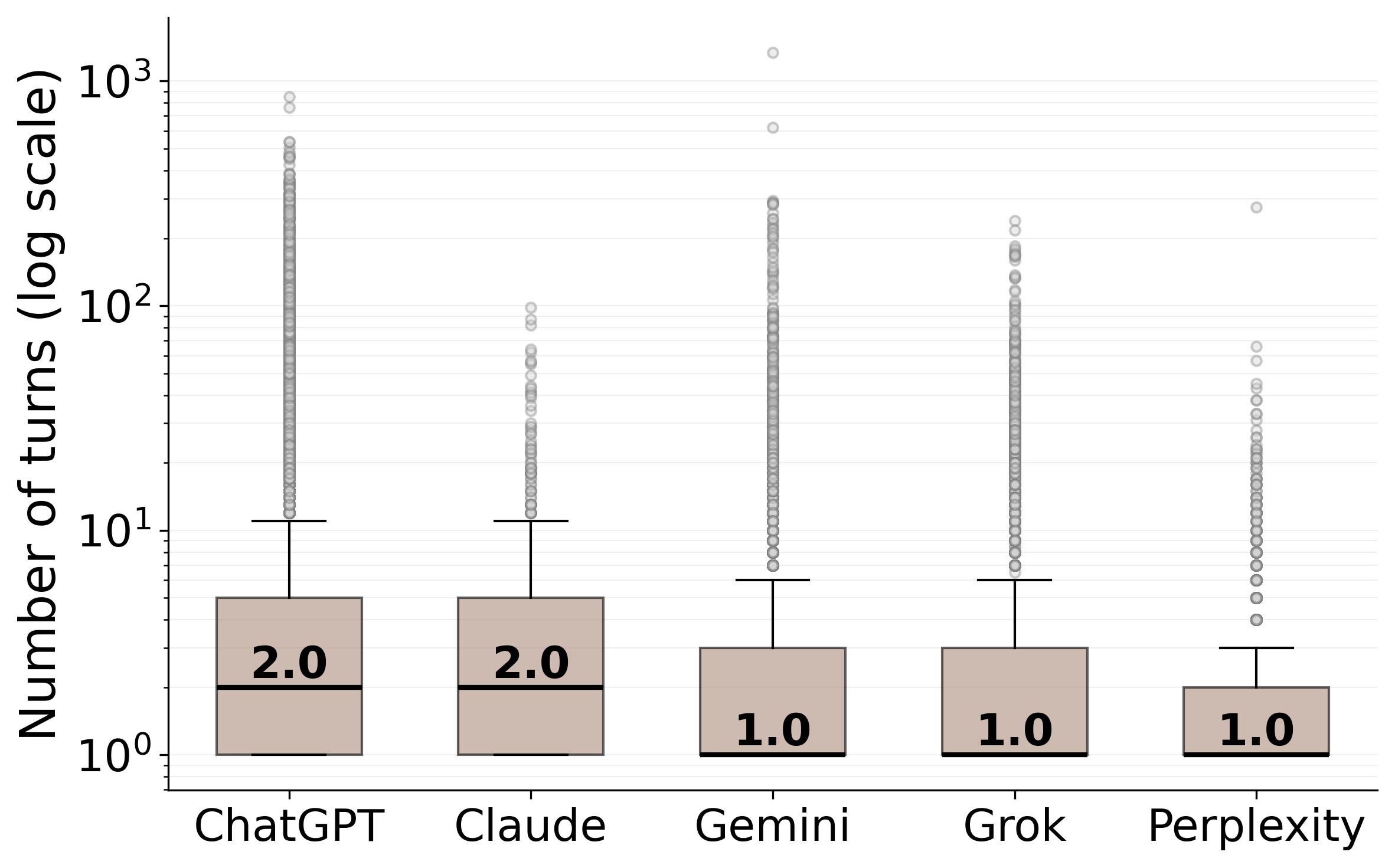}
    \caption{Turn length distribution across the five platforms in 
\textsc{ShareChat} (log scale). ChatGPT and Claude show a median of 2.0 turns, while Gemini, Grok, and Perplexity each have a median of 1.0 but with wider variability and outliers extending into the hundreds, indicating substantial diversity in conversation depth across platforms.}
    \label{fig:turns_distribution}
\end{figure}

As shown in Figure~\ref{fig:turns_distribution}, both ChatGPT and Claude show median turn lengths of 2.0, while Gemini, Grok, and Perplexity each maintain a median of 1.0 but with substantially higher variability. The presence of pronounced outliers extending into the hundreds of turns indicates substantial diversity in conversational depth, capturing both brief exchanges and extended dialogues that reflect naturalistic user interactions across platforms.  

\subsection{Toxicity Analysis Results}
\label{ssec: toxicity_results}
Leveraging the multi-label capabilities of the Detoxify model trained on the Jigsaw Toxic Comment Classification dataset, we analyzed six distinct dimensions of harmful content: \textbf{toxicity}, which is a broad category for rude or disrespectful comments, \textbf{severe toxicity}, \textbf{obscenity}, \textbf{threat}, \textbf{insult}, and \textbf{identity attack}. After filtering conversations
supported by the model, resulting in a filtered dataset of 104,107 (72.9\% of the original corpus) across all five platforms. Table~\ref{tab:toxicity_dimensions} presents the mean probability scores for each of these six dimensions by platform. 

\begin{table*}[tb!]
\centering
\footnotesize
\caption{Toxicity dimensions by platform (Detoxify-Supported Languages Only).}
\label{tab:toxicity_dimensions}
\begin{tabular}{lrrrrrrrrr}
\toprule
\textbf{Platform} & \textbf{n} & \textbf{Retention \%} & \textbf{Toxicity} & \textbf{Severe Tox.} & \textbf{Obscene} & \textbf{Threat} & \textbf{Insult} & \textbf{Identity Attack} \\
\midrule
Claude & 7,455 & 87.7 & 0.0407 & 0.0017 & 0.0297 & 0.0014 & 0.0127 & 0.0009 \\
Grok & 82,689 & 77.9 & 0.0258 & 0.0011 & 0.0115 & 0.0015 & 0.0118 & 0.0022 \\
Gemini & 52,964 & 72.8 & 0.0154 & 0.0010 & 0.0099 & 0.0017 & 0.0098 & 0.0019 \\
ChatGPT & 761,905 & 71.7 & 0.0146 & 0.0006 & 0.0070 & 0.0010 & 0.0075 & 0.0015 \\
Perplexity & 44,474 & 81.8 & 0.0088 & 0.0005 & 0.0040 & 0.0007 & 0.0051 & 0.0014 \\
\bottomrule
\end{tabular}
\end{table*}

\begin{table*}[tb!]
\centering
\footnotesize
\caption{Combined toxicity scores across all platforms (Detoxify-Supported Languages Only)}
\label{tab:toxicity_combined}
\begin{tabular}{lrrrrrrrrr}
\toprule
\textbf{Metric} & \textbf{N} & \textbf{Retention \%} & \textbf{Toxicity} & \textbf{Severe Tox.} & \textbf{Obscene} & \textbf{Threat} & \textbf{Insult} & \textbf{Identity Attack} \\
\midrule
All & 949,487 & 72.9 & 0.0211 & 0.0010 & 0.0124 & 0.0013 & 0.0094 & 0.0016 \\
llm & 471,020 & 72.9 & 0.0168 & 0.0011 & 0.0136 & 0.0006 & 0.0079 & 0.0010 \\
user & 478,467 & 72.9 & 0.0252 & 0.0009 & 0.0113 & 0.0019 & 0.0108 & 0.0021 \\
\bottomrule
\end{tabular}
\end{table*}

\begin{table*}[tb!]
\footnotesize
\centering
\caption{Conversation-level toxicity percentage comparison: Detoxify vs OpenAI moderation}
\label{tab:conversation_toxicity_comparison}
\begin{tabular}{lrrrrrrrrrrrr}
\toprule
\textbf{Role} & \multicolumn{2}{c}{\textbf{ChatGPT}} & \multicolumn{2}{c}{\textbf{Claude}} & \multicolumn{2}{c}{\textbf{Gemini}} & \multicolumn{2}{c}{\textbf{Grok}} & \multicolumn{2}{c}{\textbf{Perplexity}} & \multicolumn{2}{c}{\textbf{All Platforms}} \\
\textbf{(\%)} & \textbf{Detox} & \textbf{Open} & \textbf{Detox} & \textbf{Open} & \textbf{Detox} & \textbf{Open} & \textbf{Detox} & \textbf{Open} & \textbf{Detox} & \textbf{Open} & \textbf{Detox} & \textbf{Open} \\
\cmidrule(lr){2-3}\cmidrule(lr){4-5}\cmidrule(lr){6-7}\cmidrule(lr){8-9}\cmidrule(lr){10-11}\cmidrule(lr){12-13}
\midrule
\textbf{user} & 10.7 & 7.5 & 10.2 & 6.1 & 10.1 & 7.9 & 11.0 & 7.1 & 3.7 & 1.7 & 9.7 & 6.8 \\
\textbf{llm} & 3.1 & 6.0 & 4.8 & 5.2 & 2.9 & 6.3 & 5.2 & 8.1 & 0.7 & 2.2 & 3.0 & 5.8 \\
\textbf{All} & 11.6 & 10.2 & 11.1 & 7.7 & 10.6 & 10.3 & 12.9 & 11.0 & 3.9 & 3.0 & 10.6 & 9.4 \\
\bottomrule
\end{tabular}
\end{table*}

The results reveal substantial variation across platforms, with Claude exhibiting the highest prevalence of general toxicity with a mean score of 0.0407, followed by Grok at 0.0258, Gemini at 0.0154, ChatGPT at 0.0146, and Perplexity at 0.0088. Table~\ref{tab:toxicity_combined} further summarizes these patterns by role, demonstrating that user contributions consistently scored higher than LLM-generated content across the toxicity dimension with 0.0252 versus 0.0168, as well as in specific categories like threat and identity attack. Conversely, LLM responses showed slightly higher mean scores in the obscene dimension, measuring 0.0136 versus 0.0113.

The conversation-level results are shown in Table~\ref{tab:conversation_toxicity_comparison}. It was found that toxicity rates vary substantially by detection method, with 6.8\% to 9.7\% of conversations containing toxic user content and 3.0\% to 5.8\% containing toxic LLM responses when aggregated across platforms. The two methods diverge by role rather than uniformly: Detoxify produces higher toxicity rates on user turns at four of five platforms, whereas OpenAI Moderation produces higher rates on LLM turns at four of five. This role-dependent divergence is consistent across both turn- and conversation-level results, and likely reflects the two classifiers' differing calibration toward user-style versus assistant-style text. Furthermore, consistent with the turn-level analysis of platform trends, Perplexity exhibited the lowest prevalence of toxic conversations at roughly 3\% to 4\%, while Grok generally displayed the highest rates ranging from 11\% to 13\%. This consistency across both levels of analysis confirms that the observed differences are stable and likely stem from the distinct ways each platform is designed and used.

Conversation-level mirroring test. To accompany the within-platform mirroring claim in Section 3.2, Tables \ref{tab:mirroring_detoxify} and \ref{tab:mirroring_openai} report per-platform Spearman and Pearson correlations between per-conversation mean user toxicity and per-conversation mean LLM toxicity, computed separately for the two detection methods. Each conversation contributes one pair: the mean toxicity score across
its user turns and the mean toxicity score across its LLM turns.
Conversations missing user or LLM turns are dropped. Bonferroni correction across the five platforms is applied to all reported $p$-values. 

\begin{table}[h]
\centering
    \caption{Within-platform conversation-level mirroring test (Detoxify). Bonferroni-corrected $p$-values account for the 5 platforms tested. The conversation count $n$ is slightly smaller than the Table~\ref{table:comprehensive_comparison} totals because the test requires each conversation to contain at least one user turn and at least one LLM turn after PII filtering; conversations with one-sided turns or unmatched filenames ($\leq$0.7\% per platform) are dropped.}
    \label{tab:mirroring_detoxify}
    \begin{tabular}{lrrrrr}
    \toprule
    \textbf{Platform} & \textbf{n (convs)} & \textbf{Spearman $\rho$} & \textbf{$p_{\mathrm{raw}}$} & \textbf{$p_{\mathrm{Bonf}}$} & \textbf{Pearson $r$} \\
    \midrule
    ChatGPT    & 101,989 & 0.659 & $<10^{-300}$         & $<10^{-300}$         & 0.299 \\
    Claude     &     946 & 0.631 & $2.7\times 10^{-106}$ & $1.3\times 10^{-105}$ & 0.640 \\
    Gemini     &   7,402 & 0.646 & $<10^{-300}$         & $<10^{-300}$         & 0.329 \\
    Grok       &  14,380 & 0.565 & $<10^{-300}$         & $<10^{-300}$         & 0.465 \\
    Perplexity &  17,297 & 0.448 & $<10^{-300}$         & $<10^{-300}$         & 0.427 \\
    \bottomrule
    \end{tabular}
\end{table}

\begin{table}[h]
\centering
\caption{Within-platform conversation-level mirroring test (OpenAI Moderation).
Same construction as Table~\ref{tab:mirroring_detoxify}.}
\label{tab:mirroring_openai}
\begin{tabular}{lrrrrr}
\toprule
\textbf{Platform} & \textbf{n (convs)} & \textbf{Spearman $\rho$} & \textbf{$p_{\mathrm{raw}}$} & \textbf{$p_{\mathrm{Bonf}}$} & \textbf{Pearson $r$} \\
\midrule
ChatGPT    & 101,989 & 0.631 & $<10^{-300}$         & $<10^{-300}$         & 0.529 \\
Claude     &     946 & 0.520 & $9.0\times 10^{-67}$  & $4.5\times 10^{-66}$  & 0.797 \\
Gemini     &   7,402 & 0.606 & $<10^{-300}$         & $<10^{-300}$         & 0.566 \\
Grok       &  14,380 & 0.606 & $<10^{-300}$         & $<10^{-300}$         & 0.636 \\
Perplexity &  17,297 & 0.454 & $<10^{-300}$         & $<10^{-300}$         & 0.501 \\
\bottomrule
\end{tabular}
\end{table}

\subsection{Detailed Analysis of User Request Topic Distribution}
\label{detail_user_request_topic_distributions}

\paragraph{Number of Topics.} Our taxonomy comprises seven high-level topic categories: \textit{Multimedia}, \textit{Other/Unknown}, \textit{Practical Guidance}, \textit{Seeking Information}, \textit{Self-Expression}, \textit{Technical Help}, and \textit{Writing}. These categories provide a compact yet expressive organization of user intents, enabling consistent comparative analysis across datasets while preserving sufficient semantic granularity for downstream interpretation.

\paragraph{Discovery Prompt.} We provide below the prompt used to discover user-request topics.

\begin{tcolorbox}[
  breakable,
  colback=gray!5,
  colframe=black,
  boxrule=0.5pt,
  arc=5pt,
  left=2pt,
  right=2pt,
  top=2pt,
  bottom=2pt
]
\begin{lstlisting}[
  breaklines=true,
  breakatwhitespace=true,
  basicstyle=\scriptsize\ttfamily,
  columns=flexible
]
### SYSTEM ROLE
You are a conversation topic classifier. Output only the category name, nothing else.

### INSTRUCTIONS
1. Identify Distinct Goals: Focus on information seeking, task requests, or problem-solving goals.
2. Maintain Order: The first item in your list must correspond to the user's first real request, and so on.
3. Ignore Noise: Skip purely social turns (e.g., "Hello", "Thank you", "Okay") unless they are the only message.

You are an internal tool that classifies a message from a user to an AI chatbot, based on the context of the previous messages before it. Based on the last user message of this conversation transcript and taking into account the examples further below as guidance, please select the capability the user is clearly interested in, or `other` if it is clear but not in the list below, or `unclear` if it is hard to tell what the user even wants:
- edit_or_critique_provided_text: Improving or modifying text provided by the user.
- argument_or_summary_generation: Creating arguments or summaries on topics not provided in detail by the user.
- personal_writing_or_communication: Assisting with personal messages, emails, or social media posts.
- write_fiction: Crafting poems, stories, or fictional content.
- how_to_advice: Providing step-by-step instructions or guidance on how to perform tasks or learn new skills.
- creative_ideation: Generating ideas or suggestions for creative projects or activities.
- tutoring_or_teaching: Explaining concepts, teaching subjects, or helping the user understand educational material.
- translation: Translating text from one language to another.
- mathematical_calculation: Solving math problems, performing calculations, or working with numerical data.
- computer_programming: Writing code, debugging, explaining programming concepts, or discussing programming languages and tools.
- purchasable_products: Inquiries about products or services available for purchase.
- cooking_and_recipes: Seeking recipes, cooking instructions, or culinary advice.
- health_fitness_beauty_or_self_care: Seeking advice or information on physical health, fitness routines, beauty tips, or self-care practices.
- specific_info: Providing specific information typically found on websites, including information about well-known individuals, current events, historical events, and other facts and knowledge.
- greetings_and_chitchat: Casual conversation, small talk, or friendly interactions without a specific informational goal.
- relationships_and_personal_reflection: Discussing personal reflections or seeking advice on relationships and feelings.
- games_and_role_play: Engaging in interactive games, simulations, or imaginative role-playing scenarios.
- asking_about_the_model: Questions about the AI model's capabilities or characteristics.
- create_an_image: Requests to generate or draw new visual content based on the user's description.
- analyze_an_image: Interpreting or describing visual content provided by the user, such as photos, charts, graphs, or illustrations.
- generate_or_retrieve_other_media: Creating or finding media other than text or images, such as audio, video, or multimedia files.
- data_analysis: Performing statistical analysis, interpreting datasets, or extracting insights from data.
- unclear: If the user's intent is not clear from the conversation.
- other: If the capability requested doesn't fit any of the above categories.

Only reply with one of the capabilities above, without quotes and as presented (all lower case with underscores and spaces as shown). If the conversation has multiple distinct capabilities, choose the one that is the most relevant to the LAST message in the conversation.

Examples:
- edit_or_critique_provided_text:
"Help me improve my essay, including improving flow and correcting grammar errors."
"Please shorten this paragraph."
"Can you proofread my article for grammatical mistakes?
"Here\'s my draft speech; can you suggest enhancements?
"Stp aide moi à corriger ma dissertation."

- argument_or_summary_generation:
"Make an argument for why the national debt is important.
"Write a three-paragraph essay about Abraham Lincoln."
"Summarize the Book of Matthew."
"Provide a summary of the theory of relativity."
"R'ediger un essai sur la politique au Moyen-Orient."

personal_writing_or_communication:
"Write a nice birthday card note for my girlfriend."
"What should my speech say to Karl at his retirement party?"
"Help me write a cover letter for a job application."
"Compose an apology email to my boss."
"Aide moi à écrire une lettre à mon père."

- write_fiction:
"Write a poem about the sunset."
"Create a short story about a time-traveling astronaut."
"Make a rap in the style of Drake about the ocean."
"Escribe un cuento sobre un niño que descubre un tesoro, pero después viene un pirata."
"Compose a sonnet about time."

- how_to_advice:
"How do I turn off my screensaver?"
"My car won\'t start; what should I try?"
"Comment faire pour me connecter à mon wifi?"
"What\'s the best way to clean hardwood floors?"
"How can I replace a flat tire?"

- creative_ideation:
"What should I talk about on my future podcast episodes?"
"Give me some themes for a photography project."
"Necesito ideas para un regalo de aniversario."
"Brainstorm names for a new coffee shop."
"What are some unique app ideas for startups?"

- tutoring_or_teaching:
"How do black holes work?"
"Can you explain derivatives and integrals?"
"No entiendo la diferencia entre ser y estar."
"Explain the causes of the French Revolution."
"What is the significance of the Pythagorean theorem?"

- translation:
"How do you say Happy Birthday in Hindi?"
"Traduis Je t\'aime en anglais."
"What\'s Good morning in Japanese?"
"Translate I love coding to German."
"¿Cómo se dice Thank you en francés?"

- mathematical_calculation:
"What is 400000 divided by 23?"
"Calculate the square root of 144."
"Solve for x in the equation 2x + 5 = 15."
"What\'s the integral of sin(x)?"
"Convert 150 kilometers to miles."

- computer_programming:
"How to group by and filter for biggest groups in SQL."
"I\'m getting a TypeError in JavaScript when I try to call this function."
"Write a function to retrieve the first and last value of an array in Python."
"Escribe un programa en Python que cuente las palabras en un texto."
"Explain how inheritance works in Java."

- purchasable_products:
"iPhone 15."
"What\'s the best streaming service?"
"How much are Nikes?"
"Cuánto cuesta un Google Pixel?"
"Recommend a good laptop under $1000."

- cooking_and_recipes:
"How to cook salmon."
"Recipe for lasagna."
"Is turkey bacon halal?"
"Comment faire des crêpes?"
"Give me a step-by-step guide to make sushi."

- health_fitness_beauty_or_self_care:
"How to do my eyebrows."
"Quiero perder peso, ¿cómo empiezo?"
"What\'s a good skincare routine for oily skin?"
"How can I improve my cardio fitness?"
"Give me tips for reducing stress."

- specific_info:
"What is regenerative agriculture?"
"What\'s the name of the song that has the lyrics I was born to run?"
"Tell me about Marie Curie and her main contributions to science."
"What conflicts are happening in the Middle East right now?"
"Quelles équipes sont en finale de la ligue des champions ce mois-ci?"
"Tell me about recent breakthroughs in cancer research."

- greetings_and_chitchat:
"Ciao!"
"Hola."
"I had an awesome day today; how was yours?"
"What\'s your favorite animal?"
"Do you like ice cream?"

- relationships_and_personal_reflection:
"What should I do for my 10th anniversary?"
"I\'m feeling worried."
"My wife is mad at me, and I don\'t know what to do."
"I\'m so happy about my promotion!"
"Je sais pas ce que je fais pour que les gens me détestent. Qu\'est-ce que je fais mal?"

- games_and_role_play:
"You are a Klingon. Let\'s discuss the pros and cons of working with humans."
"I\'ll say a word, and then you say the opposite of that word!"
"You\'re the dungeon master; tell us about the mysterious cavern we encountered."
"I want you to be my AI girlfriend."
"Faisons semblant que nous sommes des astronautes. Comment on fait pour atterrir sur Mars?"

- asking_about_the_model:
"Who made you?"
"What do you know?"
"How many languages do you speak?"
"Are you an AI or a human?"
"As-tu des sentiments?"

- create_an_image:
"Draw an astronaut riding a unicorn."
"Photorealistic image of a sunset over the mountains."
"Quiero que hagas un dibujo de un conejo con una corbata."
"Generate an image of a futuristic cityscape."
"Make an illustration of a space shuttle launch."

- analyze_an_image:
"Who is in this photo?"
"What does this sign say?"
"Soy ciega, ¿puedes describirme esta foto?"
"Interpret the data shown in this chart."
"Describe the facial expressions in this photo."

- generate_or_retrieve_other_media:
"Make a YouTube video about goal kicks."
"Write PPT slides for a tax law conference."
"Create a spreadsheet for mortgage payments."
"Find me a podcast about ancient history."
"Busca un video que explique la teoría de la relatividad."

- data_analysis:
"Here\'s a spreadsheet with my expenses; tell me how much I spent on which categories."
"What\'s the mean, median, and mode of this dataset?"
"Create a CSV with the top 10 most populated countries and their populations over time. Give me the mean annual growth rate for each country."
"Perform a regression analysis on this data."
"Analyse these survey results and summarize the key findings."

- unclear:
"If there is no indication of what the user wants; usually this would be a very short prompt."

- other:
"If there is a capability requested but none of the above apply; should be pretty rare."

Okay, now your turn, taking the user conversation at the top into account: What capability are they seeking? (JUST SAY A SINGLE CATEGORY FROM THE LIST, NOTHING ELSE). If the conversation has multiple distinct capabilities, choose the one that is the most relevant to the LAST message in the conversation. The conversation to classify is:\n{conversation_text}.
\end{lstlisting}
\end{tcolorbox}

\begin{figure*}[tb!]
    \centering
    \includegraphics[width=0.75\textwidth]{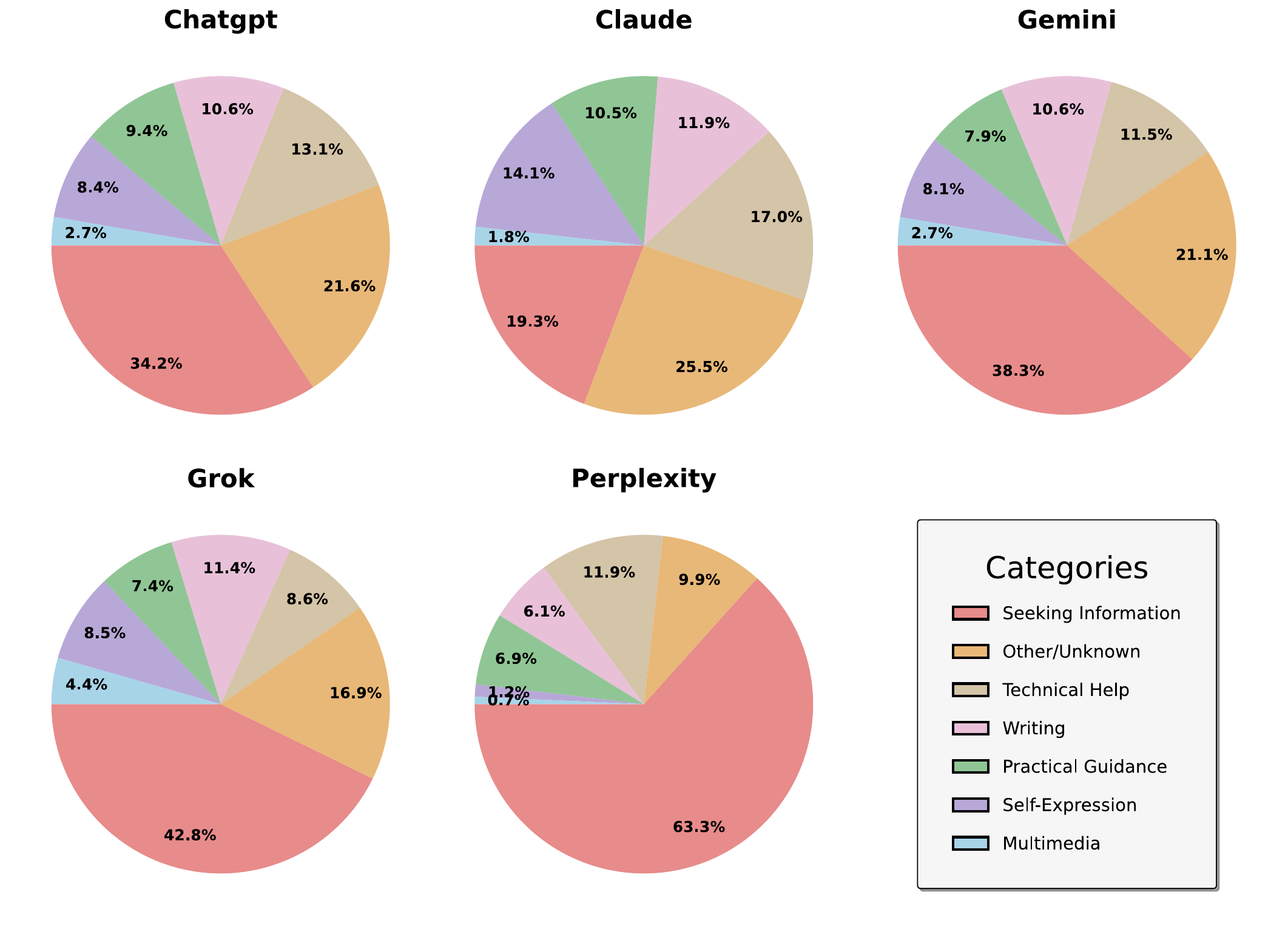}
    \caption{Per-platform topic distribution of user requests. Each pie 
chart shows the proportion of user messages classified into seven 
high-level categories. Perplexity is heavily dominated by 
\textit{Seeking Information} (63.3\%), consistent with its 
search-oriented design. Claude shows the highest share of 
\textit{Technical Help} (17.0\%) and \textit{Other/Unknown} (25.5\%), 
reflecting its positioning toward technical and reasoning tasks. 
ChatGPT, Gemini, and Grok exhibit more balanced distributions, though 
\textit{Seeking Information} remains the leading category across all 
platforms.}
    \label{fig:topic_discovery}
\end{figure*}

\paragraph{Additional Results on Topic Discovery.} 
Figure~\ref{fig:topic_discovery} presents the 
per-platform breakdown of user request topics. Several notable 
platform-specific patterns emerge beyond the aggregate distribution 
shown in the main text. Perplexity is overwhelmingly concentrated on 
\textit{Seeking Information} (63.3\%), with minimal representation of 
\textit{Self-Expression} (1.2\%) and \textit{Multimedia} (0.7\%), 
reinforcing its role as a search-oriented tool rather than a 
general-purpose assistant. Claude stands out with the highest share of 
\textit{Technical Help} (17.0\%) and the largest \textit{Other/Unknown} 
category (25.5\%), the latter likely reflecting complex, 
multi-faceted prompts that resist clean classification. ChatGPT 
(34.2\%), Gemini (38.3\%), and Grok (42.8\%) show progressively higher 
\textit{Seeking Information} shares, while \textit{Writing} remains 
relatively stable across these three platforms (10.6\%, 10.6\%, and 
11.4\% respectively). Grok exhibits the highest \textit{Multimedia} 
usage (4.4\%), potentially reflecting its integration with media-rich 
X content. These platform-level differences support the observation 
in Section~3.3 that users align their requests with perceived platform 
strengths, treating each service as a specialized tool rather than an 
interchangeable assistant.

\subsection{Conversation Completeness Analysis}
\label{conversation_completeness_prompt}

The pipeline implements a three-stage workflow adapted from the DeepEval\footnote{\url{https://github.com/confident-ai/deepeval?tab=readme-ov-file}} conversation completeness metric framework. First, we extract user intentions from each conversation by prompting Qwen3-8B \cite{yang2025qwen3} with a temperature of 0.7 to identify the distinct goals or information needs expressed in the user turns, yielding a chronologically ordered list of intentions per conversation. Second, for each identified intention, we construct an evaluation prompt that presents the full conversation context alongside the specific intention and instruct Qwen3-8B to classify whether that intention was satisfied; the model produces a categorical verdict of \textit{complete} (the assistant fully addressed the request), \textit{partial} (the assistant provided some relevant information but the goal was incompletely met), or \textit{incomplete} (the assistant failed to address the request or the conversation ended before meaningful progress). Finally, we aggregate these intention-level verdicts into a conversation-level completeness score by computing the weighted proportion of intentions, where complete verdicts contribute 1.0, partial verdicts contribute 0.5, and incomplete verdicts contribute 0.0. This three-way categorical framework allows us to capture nuanced degrees of conversational success beyond binary outcomes. We apply this pipeline uniformly across all five platforms in \textsc{ShareChat}, enabling direct cross-platform comparisons of how effectively different assistants satisfy user needs over the course of extended interactions. During manual coding, we observe that mis-classification often occurred when users pasted text, such as articles or messages, without explicit instructions or questions. In these instances, the LLMs generate responses based solely on an inferred understanding of the content.

\begin{figure}[tb!]
    \centering
    \begin{minipage}{0.48\textwidth}
        \centering
        \includegraphics[width=\linewidth]{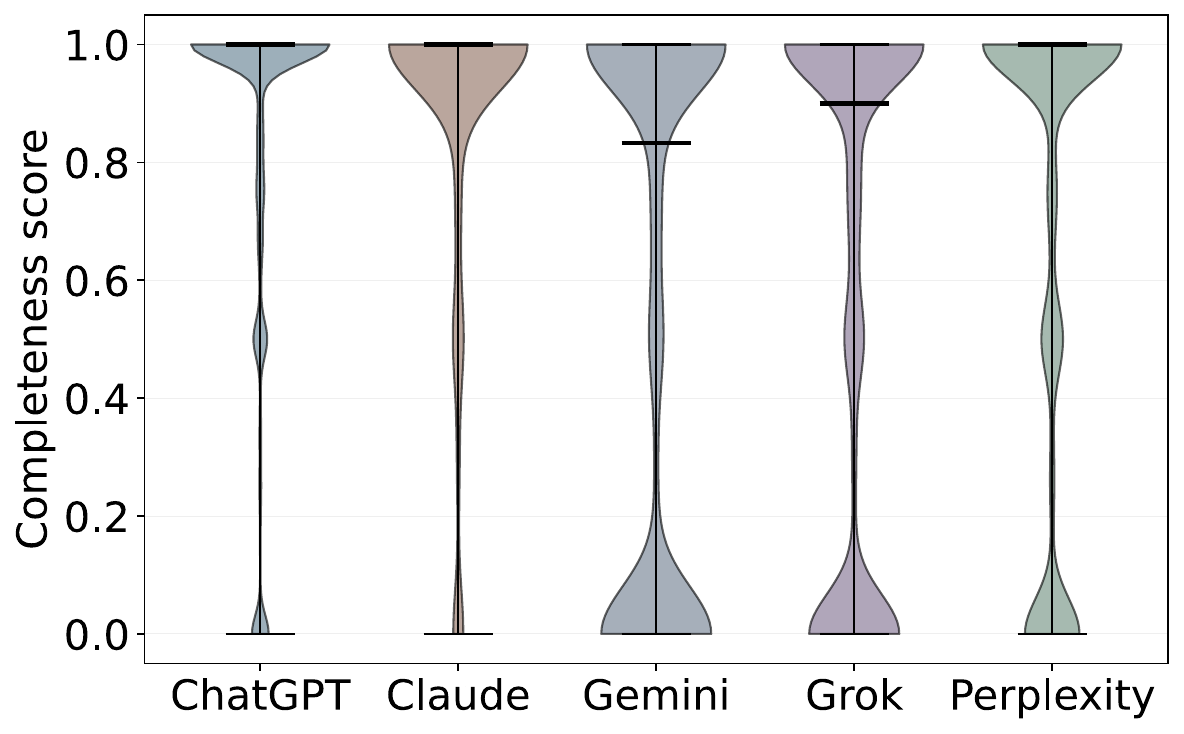}
        \caption{Conversation-level completeness score distribution across five platforms. ChatGPT, Claude, and Perplexity all achieve median scores of 1.0, indicating that most conversations fully satisfy user intentions. Gemini (median 0.833) and Grok (median 0.900) show wider distributions with more mass in the 0.2--0.6 range, reflecting greater variability in how effectively these platforms resolve multi-turn user goals.}
        \label{fig:completeness_scores}
    \end{minipage}
    \hfill 
    \begin{minipage}{0.48\textwidth}
        \centering
        \includegraphics[width=\linewidth]{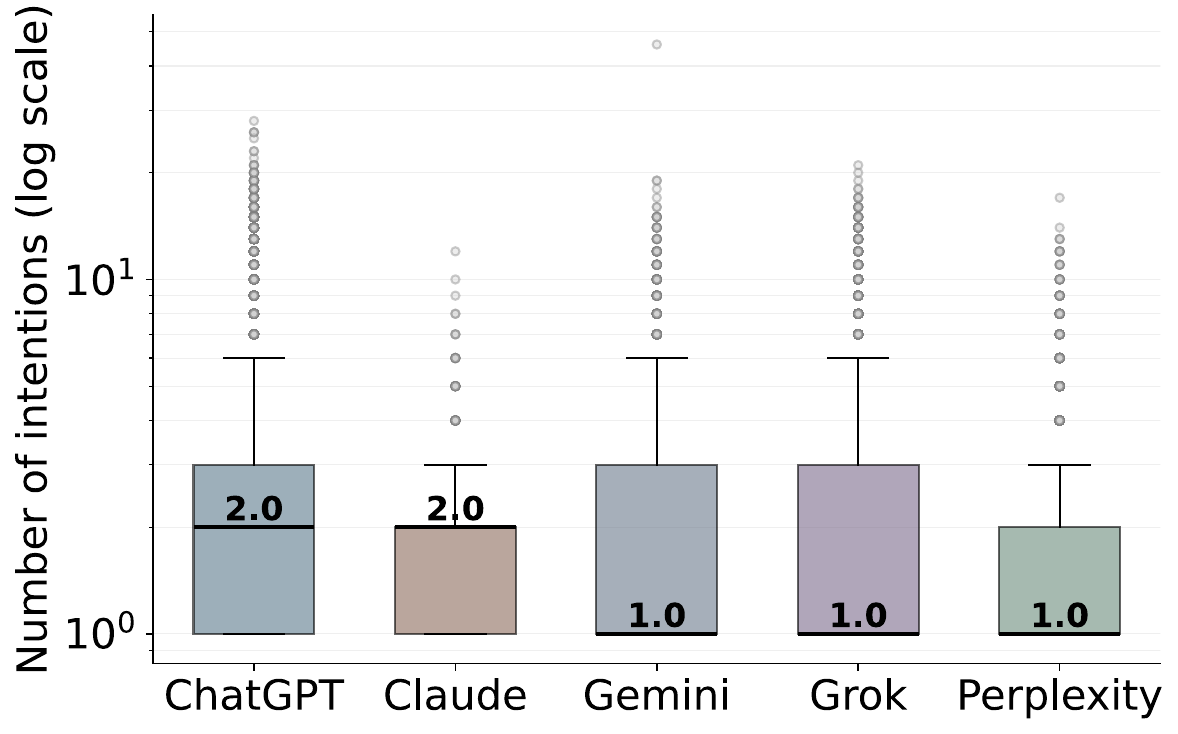}
        \caption{Number of extracted intentions per conversation (log scale), with median values labeled above each box. ChatGPT and Claude show a median of 2.0 intentions, consistent with their longer, multi-turn conversation structure, while Gemini, Grok, and Perplexity each have a median of 1.0.}
        \label{fig:completeness_intentions}
    \end{minipage}
\end{figure}

Figure~\ref{fig:completeness_scores} shows the conversation-level completeness scores, where ChatGPT, Claude, and Perplexity all achieve median scores of 1.0, while Gemini's median is 0.833 and Grok's median is 0.900. Figure~\ref{fig:completeness_intentions} shows the number of intentions across platforms.

The prompts used are as follows. To handle conversations that exceed the 40,960-token context limit of Qwen3-8B, we apply token-based truncation to assistant responses while preserving all user content, and we exclude conversations requiring severe truncation (retaining less than 80\% of content) to maintain evaluation quality.

\subsection*{Prompt A: User Intention Extraction}

\begin{tcolorbox}[
  breakable,
  colback=gray!5,
  colframe=black,
  boxrule=0.5pt,
  arc=5pt,
  left=2pt,
  right=2pt,
  top=2pt,
  bottom=2pt
]
\begin{lstlisting}[
  breaklines=true,
  breakatwhitespace=true,
  basicstyle=\scriptsize\ttfamily,
  columns=flexible
]
### SYSTEM ROLE
You are an expert Conversation Analyst.

### TASK
Extract a chronological list of **User Intentions** from the conversation log.

### INSTRUCTIONS
1. **Identify Distinct Goals:** Focus on information seeking, task requests, or problem-solving goals.
2. **Maintain Order:** The first item in your list must correspond to the user's first real request, and so on.
3. **Ignore Noise:** Skip purely social turns (e.g., "Hello", "Thank you", "Okay") unless they are the only message.

### OUTPUT FORMAT
Respond with a raw JSON object enclosed strictly within <output> tags.
The JSON must have exactly one field: "intentions" (a list of strings).

### EXAMPLE

Input Turns:
[
    {"role": "user", "content": "Hi, I need help with Python."},
    {"role": "user", "content": "How do I reverse a list?"},
    {"role": "user", "content": "Thanks. Also, what is the weather in Tokyo?"}
]

Correct Output:
<output>
{
    "intentions": [
        "User wants to know how to reverse a list in Python",
        "User wants to check the weather in Tokyo"
    ]
}
</output>

### CURRENT INPUT
Turns:
{{CONVERSATION_TEXT}}

### RESPONSE
Generate the JSON response now.
1. Start your response with the opening tag <output>.
2. Ensure the JSON is valid.
3. End with the closing tag </output>.
\end{lstlisting}
\end{tcolorbox}

\subsection*{Prompt B: Conversation Completeness Labeling}

\begin{tcolorbox}[
  breakable,
  colback=gray!5,
  colframe=black,
  boxrule=0.5pt,
  arc=5pt,
  left=2pt,
  right=2pt,
  top=2pt,
  bottom=2pt
]
\begin{lstlisting}[
  breaklines=true,
  breakatwhitespace=true,
  basicstyle=\scriptsize\ttfamily,
  columns=flexible
]
### SYSTEM ROLE
You are an expert Quality Assurance Evaluator for AI conversations.

### TASK
Determine if the specific **User Intention** was satisfied by the LLM based on the conversation history.

### CRITERIA
- **Verdict: "yes"** if:
    1. The LLM provided the correct information, code, or creative output requested.
    2. The user explicitly expressed satisfaction (e.g., "Thanks", "That works").
    3. The interaction reached a logical conclusion where the goal was met.

- **Verdict: "partial"** if:
    1. The LLM started addressing the request but the conversation ended before completion.
    2. The LLM provided some relevant information but missed key aspects of the request.
    3. The LLM gave a partial solution that requires additional steps the user would need to complete.
    4. The user asked follow-up questions indicating partial understanding/satisfaction.

- **Verdict: "no"** if:
    1. The LLM refused the request (unless it was a safety violation).
    2. The LLM completely misunderstood the request or provided irrelevant information.
    3. The user expressed frustration or repeatedly asked the same thing without progress.
    4. The LLM asked for clarification but the conversation ended before any attempt to help.

### OUTPUT FORMAT
Respond with a raw JSON object enclosed strictly within <output> tags.
The JSON must have these fields:
- "intention": (repeat the intention text)
- "verdict": (value must be "yes", "partial", or "no")

### EXAMPLE

Intention: "User wants to learn about Python decorators"
Turns: [
    {"role": "user", "content": "Can you explain Python decorators?"},
    {"role": "assistant", "content": "Sure! Decorators are functions that modify other functions. They use @ syntax. Would you like to see a code example?"}
]

Correct Output:
<output>
{"intention": "User wants to learn about Python decorators", "verdict": "yes"}
</output>

### CURRENT INPUT
Intention: {{USER_INTENTION}}

Turns:
{{CONVERSATION_TEXT}}

### RESPONSE
Generate the JSON response now.
1. Start your response with the opening tag <output>.
2. Ensure the JSON is valid.
3. End with the closing tag </output>.
\end{lstlisting}
\end{tcolorbox}


\subsection{Additional Source Grounding Analysis}
\label{appendix:response_source_analysis}

\begin{figure}[htb]
    \centering
    \includegraphics[width=0.7\textwidth]{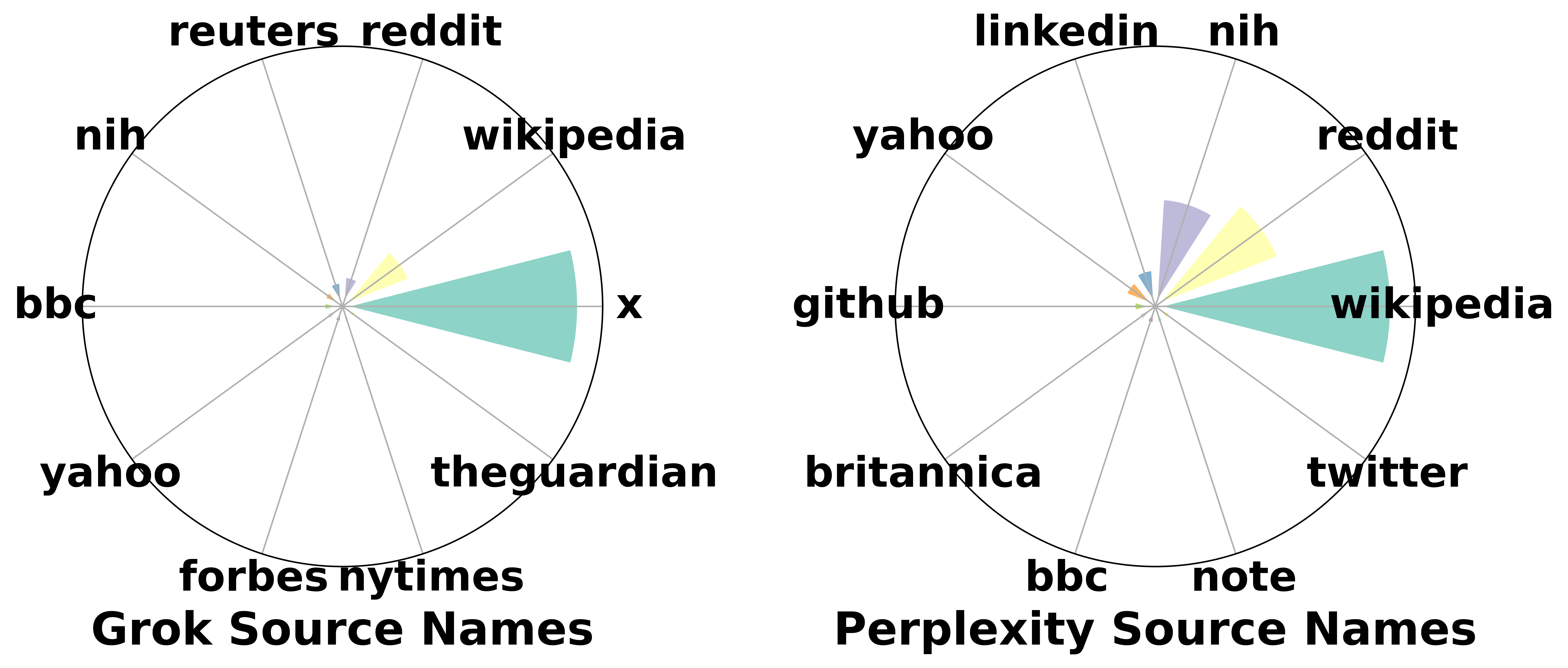}
    \caption{Response source top frequency rose graph presents the rose plots of the top 10 most frequent source domains, highlighting X as Grok’s dominant source and Wikipedia as Perplexity’s leading source.}
    \label{fig:response_source_analysis_rose_graph}
\end{figure}

Figure~\ref{fig:response_source_analysis_rose_graph} presents rose 
plots of the top 10 most frequently cited source domains for Grok and 
Perplexity. For Grok, X dominates with 40,624 references, followed by 
Wikipedia (12,507) and a cluster of major news outlets including BBC, 
The Guardian, Yahoo, and the New York Times. For Perplexity, the 
distribution is notably more balanced: Wikipedia leads with 2,919 
citations, followed by Reddit (1,642), NIH (1,339), and a mix of 
encyclopedic (Britannica), professional (GitHub, LinkedIn), and news 
sources (BBC, Twitter/X). The contrast highlights Grok's reliance on 
a single social media platform versus Perplexity's diversified 
retrieval across authoritative and community-driven sources.

\begin{figure}[tb!]
    \centering
    \begin{subfigure}[b]{0.48\textwidth}
        \centering
        \includegraphics[width=\textwidth]{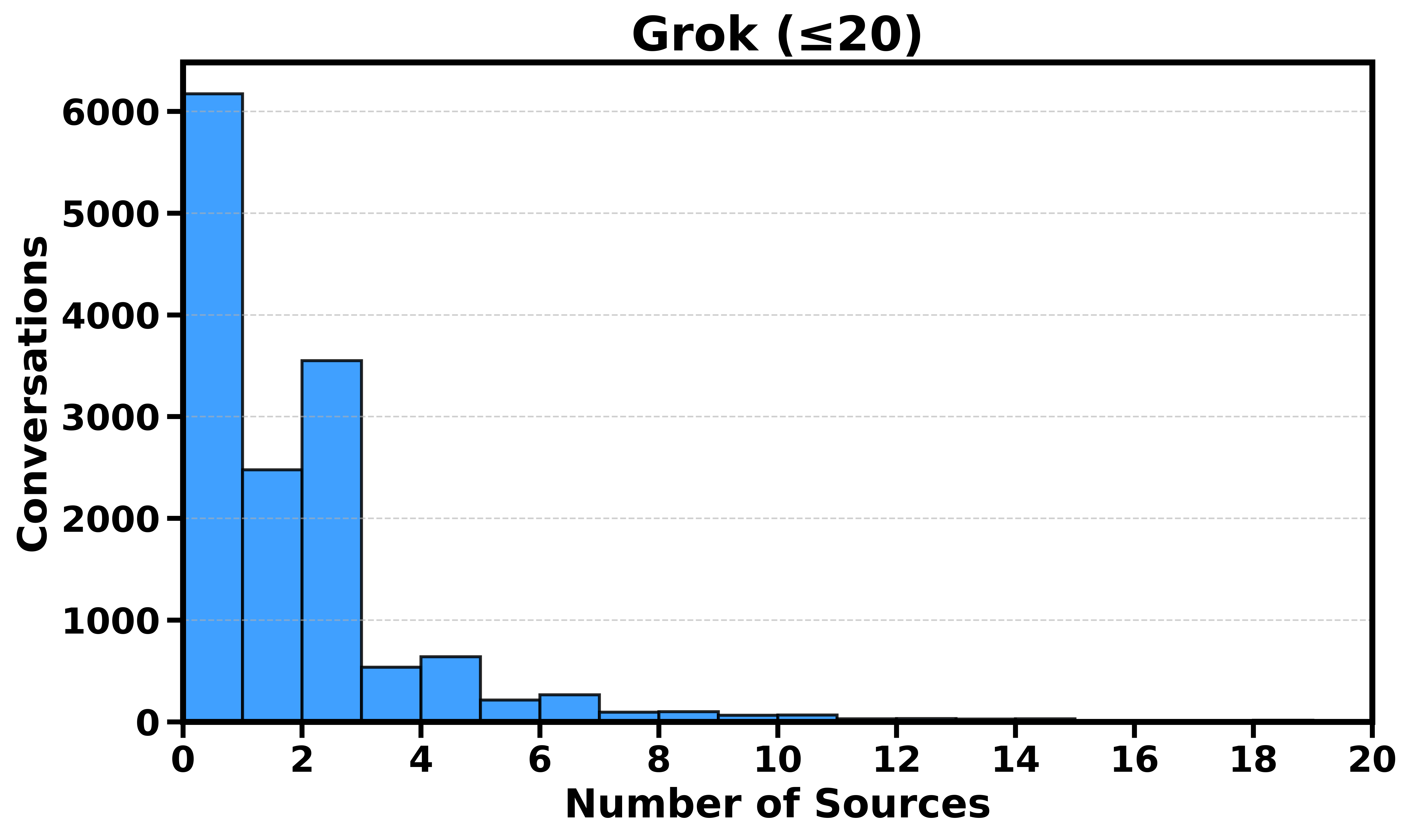}
        \caption{Grok}
        \label{fig:grok_source_dist}
    \end{subfigure}
    \begin{subfigure}[b]{0.48\textwidth}
        \centering
        \includegraphics[width=\textwidth]{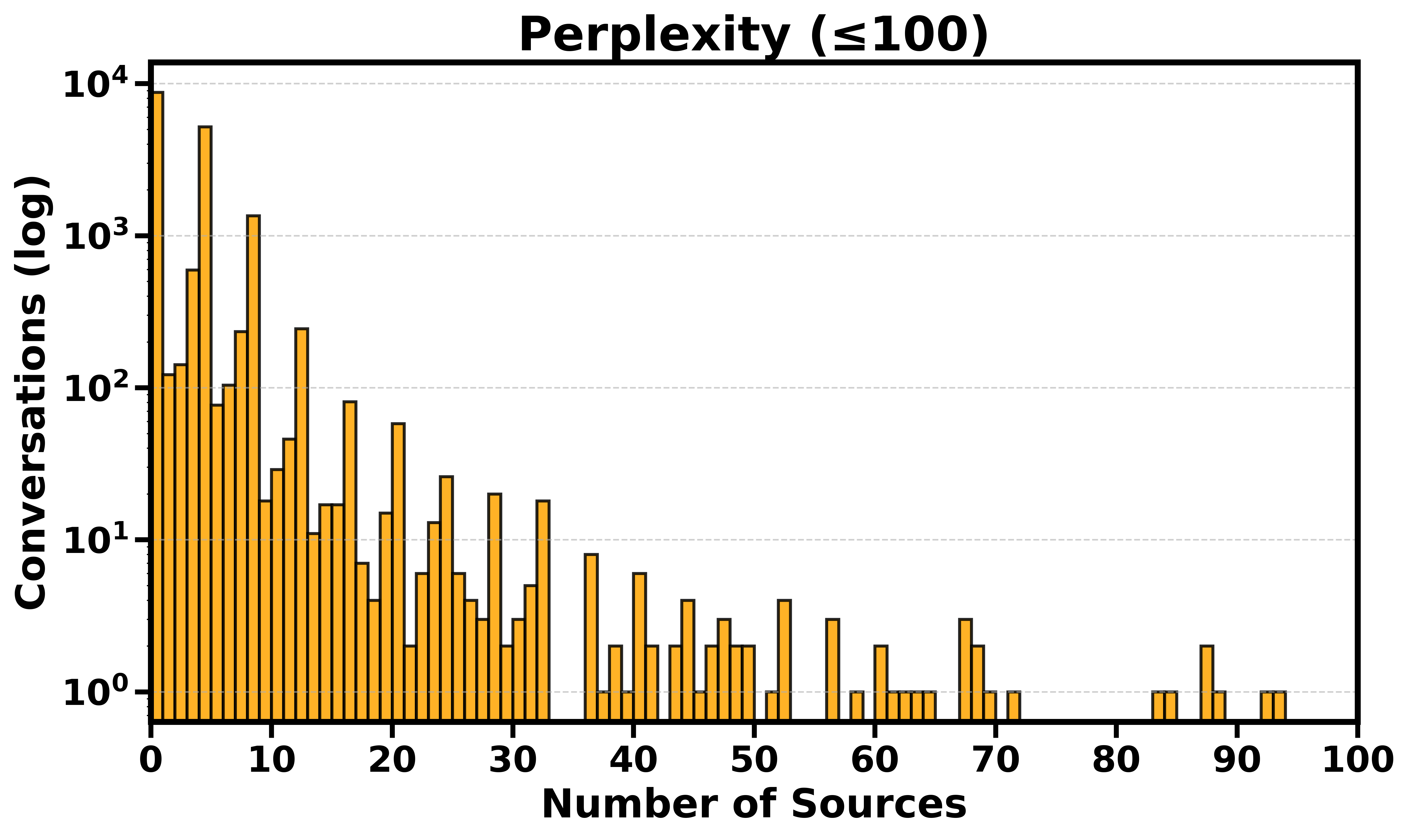}
        \caption{Perplexity}
        \label{fig:perplexity_source_dist}
    \end{subfigure}
    \caption{Distribution of source citations per conversation for Grok and Perplexity. Extreme tails are omitted for visual clarity; the maximum observed source count is 83 for Grok and 1,059 for Perplexity.}
    \label{fig:response_source_analysis_count_distribution}
\end{figure}

To further understand retrieval intensity, we analyze the distribution 
of source counts per conversation in 
Figure~\ref{fig:response_source_analysis_count_distribution}. Grok 
conversations typically cite very few sources, with the majority 
drawing on fewer than 5 and a maximum of 83. Perplexity, by contrast, 
exhibits a heavy-tailed distribution on a logarithmic scale: while 
most conversations cite between 1 and 20 sources, a substantial 
number integrate 30--60 sources, with the maximum reaching 1,059. 
This order-of-magnitude difference in retrieval depth reflects the 
two platforms' distinct design philosophies: Grok prioritizes 
targeted, real-time social signals, while Perplexity supports 
extensive, research-oriented information gathering across a broad 
range of domains.

\subsection{Additional Temporal Analysis}
\label{ssec: temporal}

\begin{figure}[t]   
    \centering 
    \includegraphics[width=0.6\linewidth]{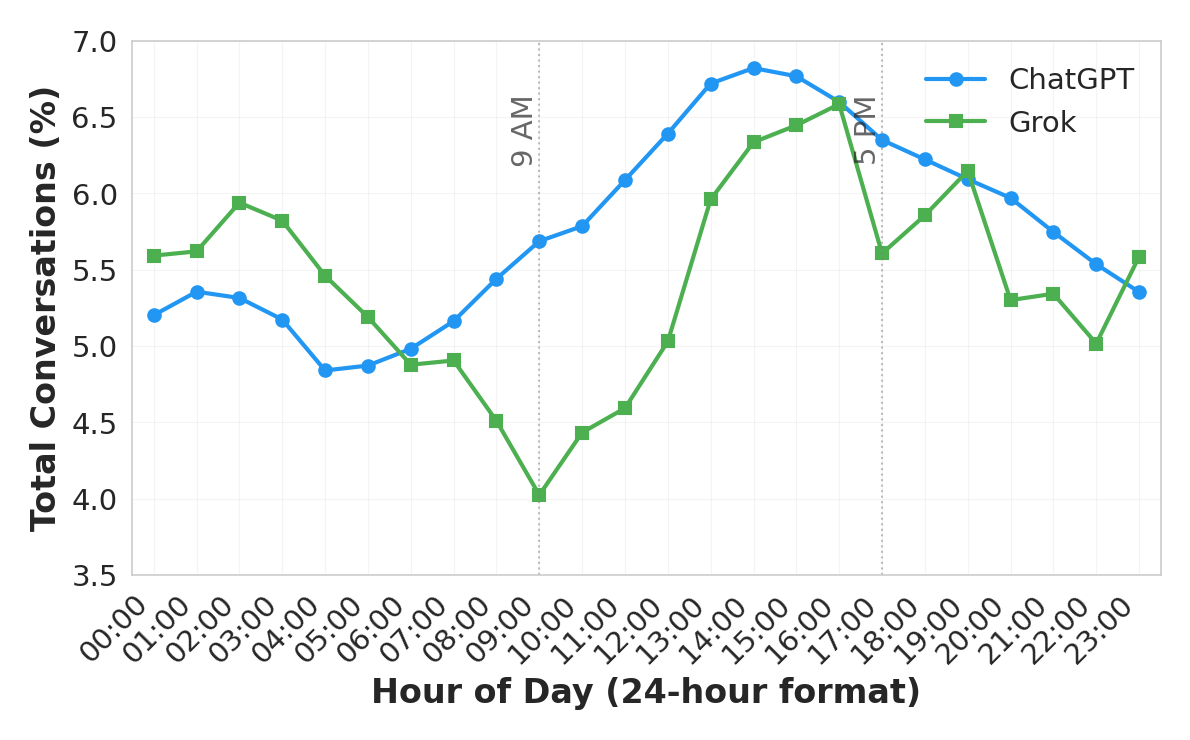}
    \label{fig:hourly_activity_line}
    \caption{Normalized hourly activity distribution for ChatGPT and Grok, with timestamps adjusted to users' local time zones. Both platforms exhibit a trough during early morning hours (2:00--6:00) and peak activity in the late morning to early evening. ChatGPT activity peaks around 9~AM and again at 5~PM, while Grok shows a similar but slightly flatter diurnal pattern. The overall similarity suggests that usage rhythms are driven primarily by daily routines rather than platform-specific factors.}
    \label{fig:combined_activity_and_response}
\end{figure}

\begin{figure}[tb]
    \centering
    \begin{subfigure}{0.48\columnwidth} 
        \centering
        \includegraphics[width=\linewidth]{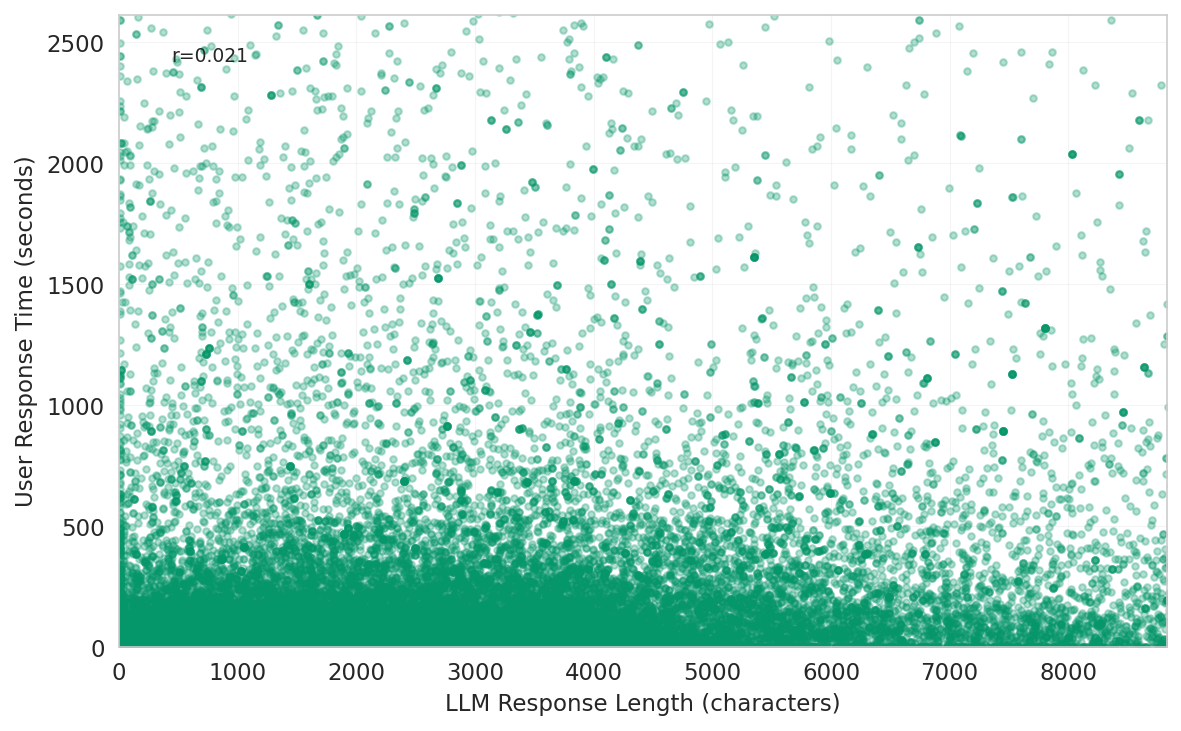} 
        \caption{Grok}
        \label{fig:app_grok_length_vs_response}
    \end{subfigure}
    \hfill 
    \begin{subfigure}{0.48\columnwidth} 
        \centering
        \includegraphics[width=\linewidth]{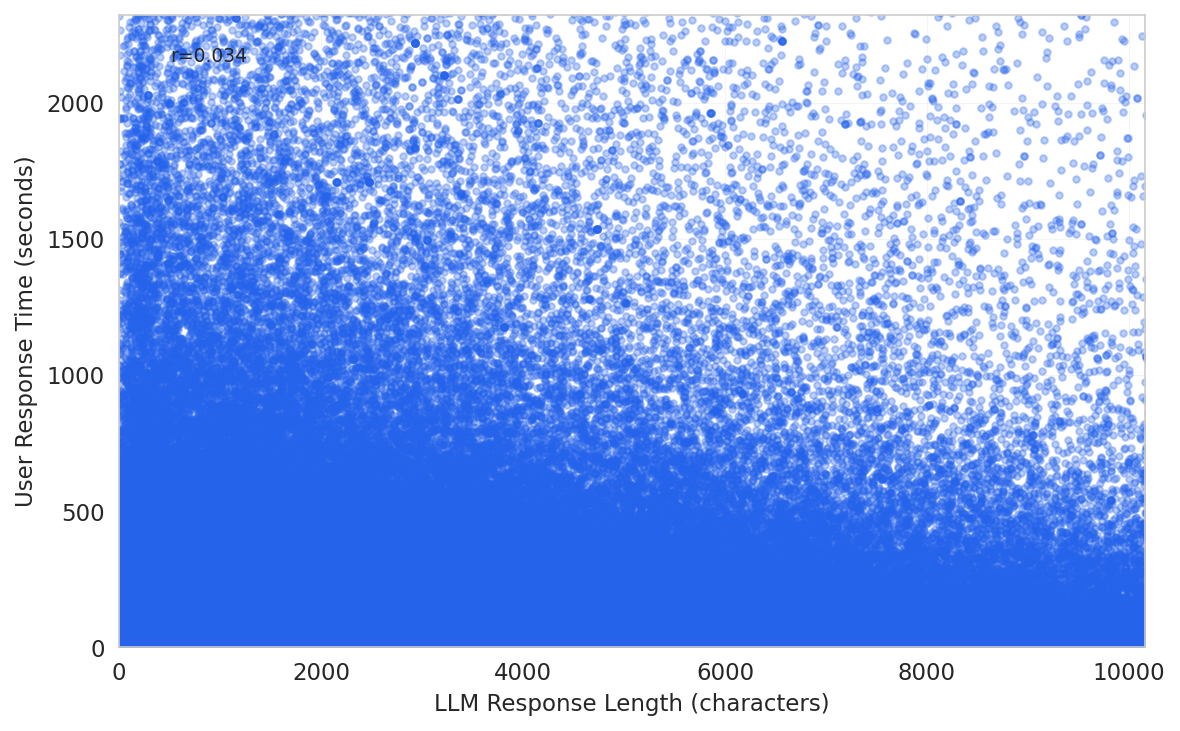} 
        \caption{ChatGPT}
        \label{fig:app_chatgpt_length_vs_response}
    \end{subfigure}
    
    \caption{Raw scatter plots of user response time (seconds) versus LLM response length (characters) for (a)~Grok and (b)~ChatGPT. Both 
platforms show high variance in user response time across all response 
lengths, with Pearson correlations near zero (Grok: $r = 0.021$; 
ChatGPT: $r = 0.034$), indicating that model verbosity alone is a 
poor predictor of individual-level user dwell time.}
    \label{fig:app_length_vs_response}
\end{figure}

We expand on the temporal dynamics of the dataset by examining both the diurnal rhythms of user activity and the granular relationship between content length and reading time. First, to understand when users are most active, Figure~\ref{fig:combined_activity_and_response} presents the normalized hourly activity for each platform, with timestamps adjusted to the users' local time. 
Second, we investigate whether longer model outputs consistently drive longer user engagement. 
Figure~\ref{fig:app_length_vs_response} plots the raw distribution of user response times against LLM response length. The scatter plot reveals a high degree of variance, with Pearson correlations near zero for both platforms (ChatGPT: $r = 0.034$; Grok: $r = 0.021$). This indicates that while aggregate trends exist, model verbosity is a poor predictor of dwell time.

\subsection{PII Removal Validation and Data Filtering}
\label{sssec:pii}

To validate the efficacy of our de-identification pipeline, we employed GPT-OSS-120B, accessed through the Jetstream Cloud API\footnote{https://docs.jetstream-cloud.org/inference-service/overview/}, as an automated auditor to detect any residual personally identifiable information in the anonymized text. The model was configured with medium reasoning effort to optimize for computational efficiency while maintaining evaluation quality. The quantitative results of this automated audit are presented in Table~\ref{tab:pii_audit}.

\begin{table}[htb]
    \centering
    \small
    \caption{Automated PII removal audit results. ``Success Rate'' indicates the percentage of records classified as clean by the LLM evaluator. "Flagged" denotes the number of records the evaluator suspected of containing residual PII.}
    \label{tab:pii_audit}
    \begin{tabular}{lrrr}
        \toprule
        \textbf{Platform} & \textbf{Success Rate} & \textbf{Flagged} & \textbf{Total} \\
        \midrule
        ChatGPT  & 95.20\% & 51041   & 1062949 
        \\
        Perplexity  & 94.42\% & 2,899 & 54,355  \\
        Grok        & 94.15\% & 6,010 & 106,168 \\
        Gemini      & 95.43\% & 3,302 & 72,746  \\
        Claude      & 97.01\% & 252   & 8,504   \\ 
        \bottomrule
    \end{tabular}
\end{table}

To further verify these automated findings, we conducted a manual review of 50 randomly sampled entries (228 turns) that the evaluator flagged as containing PII (i.e., "Positive" labels). This human verification revealed a true positive rate of only 3.7\%, indicating that the vast majority of flagged instances were false positives and that the pipeline is highly effective at removing sensitive data.


To ensure that contextual-PII review covers the bulk of every released
conversation, we apply a conversation-level majority-language filter:
a conversation is retained in the release only if more than 50\% of
its messages are in one of the 11 languages supported by Presidio's
NER models (English, Spanish, German, French, Italian, Portuguese,
Dutch, Chinese, Japanese, Russian, and Hebrew). Conversations whose
majority language falls outside this set are dropped entirely; their
original source URLs remain available to future researchers wishing
to apply language-specific PII tooling. Within retained conversations,
minority-language messages are preserved, since the conversation's
surrounding context has already received contextual-PII review. All
redaction spans returned by Presidio, irrespective of which
recognizer fired or which entity type was assigned, are post-processed
into a single \texttt{<REDACTED>} placeholder for a uniform release
schema. The resulting per-platform retention rates are reported in
Table~\ref{tab:final_dataset_stats}.

\begin{table}[htb]
    \centering
\caption{Final dataset statistics after applying language and PII 
safety filtering. Counts represent individual \textit{messages} 
(user and assistant turns combined), not conversations. ``Original'' 
denotes the total number of messages processed through the 
de-identification pipeline, ``Removed'' indicates messages excluded 
due to unsupported languages or residual PII concerns, and ``\% 
Kept'' reflects the retention rate after filtering.}
    \label{tab:final_dataset_stats}
    \small
    \begin{tabular}{lrrr}
        \toprule
        \textbf{Platform} & \textbf{Original} & \textbf{Removed} & \textbf{\% Kept} \\
        \midrule
        ChatGPT     & 1,062,949    & 86,571 & 91.9\% \\
        Perplexity  & 54,355        & 5,754  & 89.4\% \\
        Grok        & 106,168       & 7,280  & 93.1\% \\
        Gemini      & 72,746       & 5,078  & 93.0\% \\
        Claude      & 8,504         & 140    & 98.4\% \\
        \midrule
        \textbf{Total} & \textbf{1,304,722} & \textbf{104,823} & \textbf{92.0\%} \\
        \bottomrule
    \end{tabular}
\end{table}

\clearpage
\newpage

\end{document}